\documentclass{article}

\PassOptionsToPackage{numbers, compress}{natbib}

\usepackage[preprint]{neurips_2025}

\usepackage[utf8]{inputenc} 
\usepackage[T1]{fontenc}    
\usepackage{hyperref}       
\usepackage{url}            
\usepackage{booktabs}       
\usepackage{amsfonts}       
\usepackage{nicefrac}       
\usepackage{microtype}      
\usepackage{xcolor}         
\usepackage{graphicx}
\usepackage{caption}
\usepackage{makecell}
\usepackage{comment}
\usepackage{tikz}
\usepackage{pgfplots}
\usepackage{subcaption}
\usepackage{xspace}
\usepackage{tabularx}
\usepackage{enumitem}
\usepackage{multirow}
\usepackage{makecell}
\usepackage{amssymb}
\usepackage{pifont}
\usepackage{cleveref}
\usepackage{threeparttable}

\usepackage{rotating}   
\usepackage{tabularx}   
\usepackage{booktabs}   
\usepackage{multirow}   
\usepackage{graphicx}   

\newcommand{\cmark}{\ding{51}}%
\newcommand{\xmark}{\ding{55}}%

\newcommand{\blueContent}[1]{{\color{blue}#1}}

\makeatletter
\DeclareRobustCommand\onedot{\futurelet\@let@token\@onedot}
\def\@onedot{\ifx\@let@token.\else.\null\fi\xspace}

\def\eg{\emph{e.g}\onedot}

\def\etc{\emph{etc}\onedot} 
 
\def\etal{\emph{et al}\onedot}
\makeatother

\title{Many-for-Many: Unify the Training of Multiple Video and Image Generation and Manipulation Tasks}

\author{
    Ruibin Li\textsuperscript{1,2}\thanks{Equal contribution.}\quad
    Tao Yang\textsuperscript{1}\footnotemark[1]\;\thanks{Corresponding author.}\quad
    Yangming Shi\textsuperscript{1}\quad
    \textbf{Weiguo Feng\textsuperscript{1}}\quad \\
    \textbf{Shilei Wen\textsuperscript{1}}\quad
    \textbf{Bingyue Peng\textsuperscript{1}}\quad
    \textbf{Lei Zhang\textsuperscript{2}\footnotemark[2]} \\
    \textsuperscript{1}ByteDance\quad
    \textsuperscript{2}The Hong Kong Polytechnic University\quad
    \\
}

\pgfplotsset{compat=1.18}

\begin{document}

\maketitle

\begin{abstract}
  Diffusion models have shown impressive performance in many visual generation and manipulation tasks. Many existing methods focus on training a model for a specific task, especially, text-to-video (T2V) generation, while many other works focus on finetuning the pretrained T2V model for image-to-video (I2V), video-to-video (V2V), image and video manipulation tasks, \etc. However, training a strong T2V foundation model requires a large amount of high-quality annotations, which is very costly. In addition, many existing models can perform only one or several tasks. In this work, we introduce a unified framework, namely \textit{many-for-many}, which leverages the available training data from many different visual generation and manipulation tasks to train a single model for those different tasks. Specifically, we design a lightweight adapter to unify the different conditions in different tasks, then employ a joint image-video learning strategy to progressively train the model from scratch. Our joint learning leads to a unified visual generation and manipulation model with improved video generation performance. In addition, we introduce depth maps as a condition to help our model better perceive the 3D space in visual generation. Two versions of our model are trained with different model sizes (8B and 2B), each of which can perform more than 10 different tasks. In particular, our 8B model demonstrates highly competitive performance in video generation tasks compared to open-source and even commercial engines. Our models are available at \blueContent{\href{https://huggingface.co/LetsThink/MfM-Pipeline-8B}{MfM-Pipeline-8B}, \href{https://huggingface.co/LetsThink/MfM-Pipeline-2B}{MfM-Pipeline-2B}} and source codes are available at \blueContent{\href{https://github.com/leeruibin/MfM.git}{https://github.com/leeruibin/MfM.git}}.
\end{abstract}

\section{Introduction}
Visual data generation has a wide range of applications in industry and our daily lives, such as video games \cite{valevski2024game}, advertising \cite{zhang2024virbo}, media content creation \cite{polyak2025moviegen}, \etc. Along with the great success of text-to-image (T2I) generation models \cite{ramesh2021dalle,rombach2021latent,podell2023sdxl,esser2024sd3}, video generation techniques \cite{openai2024sora,yang2024cogvideox,polyak2025moviegen,ma2025stepvideot2v,kong2024hunyuanvideo,chen2025goku,wan2.1} have recently witnessed significant progress driven by the rapid development of diffusion models (DMs) \cite{ho2020ddpm,rombach2021latent,peebles2022dit,lipman2023fm}. Current research is preliminarily focused on text-to-video (T2V) generation. Early attempts \cite{guo2023animatediff,blattmann2023videoldm,blattmann2023svd} are often built on pre-trained T2I models such as Stable Diffusion (SD) \cite{rombach2021latent} by encoding motion dynamics into latent codes \cite{khachatryan2023text2video-zero} or inserting additional temporal layers \cite{guo2023animatediff,blattmann2023videoldm,blattmann2023svd}. Despite significant advancements, these methods tend to produce unnatural motions and are limited by the small number of generated frames.

Recently, diffusion transformers (DiT) \cite{peebles2022dit,esser2024sd3,yang2024cogvideox} have been widely adopted in numerous image and video generation methods \cite{esser2024sd3,flux,openai2024sora,wan2.1} due to their excellent scalability. In particular, SORA \cite{openai2024sora} demonstrates remarkable performance in the creation of highly realistic videos, sparking widespread interest in the community and inspiring many subsequent T2V works  \cite{yang2024cogvideox,ma2025stepvideot2v,kong2024hunyuanvideo,wan2.1,runway,kling}.
For example, trained on web-scale datasets, the open-source models CogVideoX \cite{yang2024cogvideox}, HunyuanVideo \cite{kong2024hunyuanvideo} and Wan2.1 \cite{wan2.1} have attracted significant attention. The commercial models Runway \cite{runway} and Kling \cite{kling} have demonstrated impressive performance in practical use. Other video generation tasks, such as image-to-video (I2V) and video-to-video (V2V), are commonly regarded as downstream problems of T2V. By fine-tuning pre-trained T2V models with relatively small resources \cite{yang2024cogvideox,kong2024hunyuanvideo,wan2.1}, various models tailored to different tasks can be obtained, including I2V \cite{tian2025edg,wan2.1}, video super-resolution \cite{xie2025star}, reference-to-video \cite{liu2025phantom,jiang2025vace}, \etc. 

In this work, we aim to economically train a model from scratch, which can, however, perform a number of visual generation and manipulation tasks effectively, including T2V, I2V, V2V, \etc. To this end, we introduce a simple yet effective framework, called \textit{Many-for-Many} (MfM in short), to unify the training of different tasks. The key difference between the various visual generation/manipulation tasks lies in their varying conditions. Therefore, we propose to standardize the conditions using a lightweight adapter, thereby enabling multi-task joint training. Adhering to the foundation model's training recipe, we progressively update our MfM model from a low resolution to higher resolutions. Specifically, we employ a joint image-video learning strategy, which equips our model with capabilities for both image generation and manipulation. An advantage of our MfM training framework is that the many data that cannot be used to train T2V models in previous methods now can be used to train our unified model. Therefore, MfM learning not only leads to a unified model but also enhances video generation performance. Additionally, we incorporate depth maps as a condition to improve our model's understanding of 3D space in the real world. 

\begin{figure}[t!]
\includegraphics[width=\textwidth]{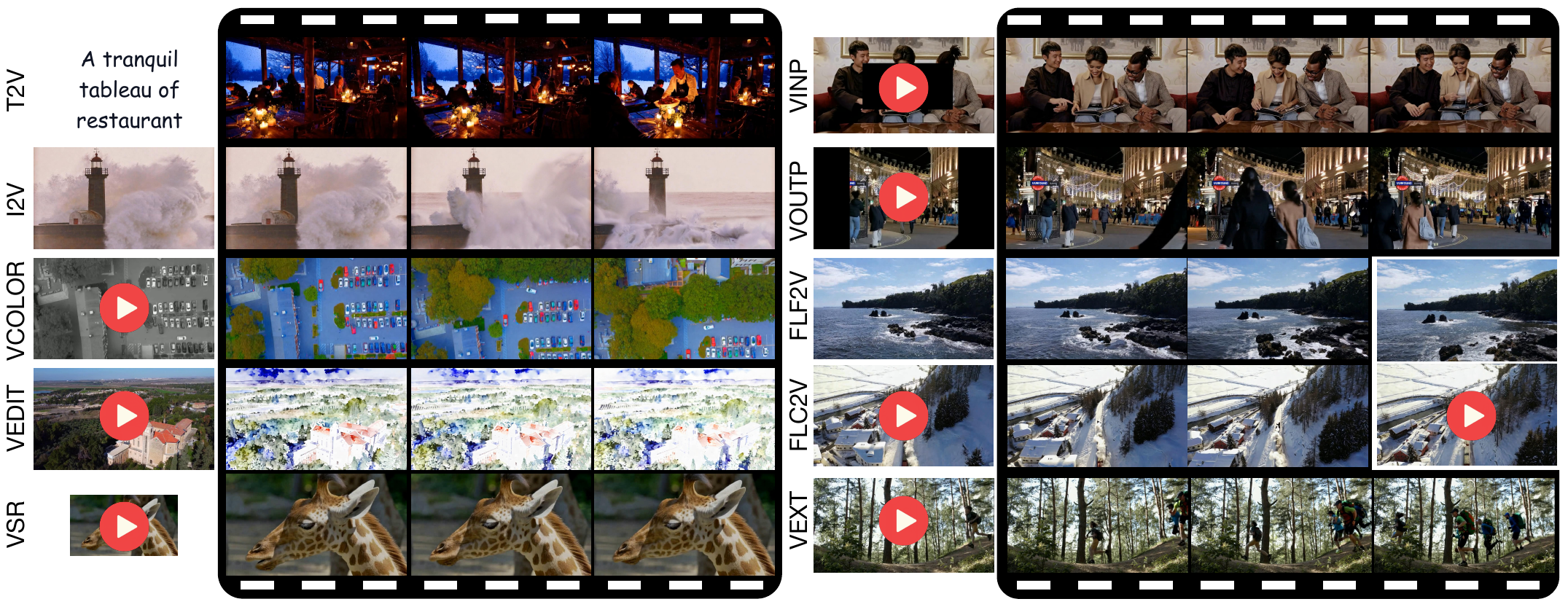}
\vspace{-3mm}
\caption{Examples of MfM on typical video generation and manipulation tasks. Generated frames are highlighted within black boxes. Note that MfM uses a single model to perform these tasks.}
\centering
\label{fig:visual}
\vspace{-4mm}
\end{figure}

Two versions (2B and 8B) of our MfM model are trained. As shown in Table~\ref{tab:modellist}, our model can perform more than 10 different visual generation and manipulation tasks. Figure~\ref{fig:visual} illustrates some examples of MfM tasks. Extensive experiments are performed to demonstrate the effectiveness and flexibility of our MfM model. In particular, our 8B model achieves highly competitive performance in the challenging T2V and I2V tasks by using only 10\% of the training data used in state-of-the-art open-source T2V foundation models \cite{yang2024cogvideox,kong2024hunyuanvideo,wan2.1}.

\begin{table}[t!]
\centering
\caption{The size and supported tasks of the current main open-source video foundation models.}
\vspace*{1mm}
\normalsize{
\resizebox{0.9\textwidth}{!}{\begin{tabular}{|c|c|c|c|c|c|}
\hline
\multirow{2}{*}{Model} & \multirow{2}{*}{Size} & \multicolumn{2}{|c|}{Training Data} & \multirow{2}{*}{Supported Tasks} & \multirow{2}{*}{Unified Training} \\
\cline{3-4}
& & Video & Image & & \\
\hline
CogVideoX \protect\cite{yang2024cogvideox} & 28\&5B & unkown & unkown & T2V, I2V & \xmark \\
\hline
MovieGen \protect\cite{polyak2025moviegen} & 30B & 100M  & 1B & T2V, Peronalized T2V (PT2V) & \xmark \\
\hline
StepVideo \protect\cite{ma2025stepvideot2v} & 30B & 2B & 3.8B & T2V, I2V & \xmark \\
\hline
HunyuanVideo \protect\cite{kong2024hunyuanvideo} & 13B & $\mathcal{O}(100)$M & $\mathcal{O}(1)$B & T2V, I2V & \xmark \\
\hline
Wan2.1 \protect\cite{wan2.1} & 1.3B\&14B & 1.5B & 10B & T2V, I2V & \xmark \\
\hline
MfM & 2B\&8B & 120M & 160M & \makecell{T2V, I2V, video extension, \\ FLF2V, FLC2V, video manipulation, etc.} & \cmark \\
\hline
\end{tabular}}}
\label{tab:modellist}
\vspace{-2mm}
\end{table}

\section{Related Work}
\textbf{Diffusion Models for Visual Generation}.
Since the seminal work of denoising diffusion probabilistic model (DDPM) \cite{ho2020ddpm}, remarkable progress has been achieved in training diffusion models (DMs) for image and video generation \cite{rombach2021latent,esser2024sd3,blattmann2023svd,openai2024sora,kong2024hunyuanvideo}. In particular, Rombach \etal \cite{rombach2021latent} proposed to train DMs in latent space, achieving impressive image generation results with significantly reduced computational costs. The development of Stable Diffusion (SD) \cite{rombach2021latent} has sparked a surge of research in text-to-image (T2I) generation \cite{podell2023sdxl,zhang2023controlnet,ruiz2023dreambooth}. SDXL \cite{podell2023sdxl} expands SD by using a larger model and more sophisticated architecture design. With the advancement in Diffusion Transformer (DiT) \cite{peebles2022dit} and Flow Matching (FM) \cite{lipman2023fm}, Esser et al. \cite{esser2024sd3} proposed MMDiT to scale up the pre-training of T2I models, resulting in SD3 and Flux \cite{flux}, which show new state-of-the-art T2I performance.

In terms of T2V generation, early efforts often leverage pre-trained T2I models and fine-tune them to learn motion dynamics \cite{guo2023animatediff,blattmann2023svd,chen2024videocrafter2}, which are, however, limited in both motion naturalness and video frames. The great success of SORA \cite{openai2024sora} in generating highly realistic long videos has inspired numerous commercial \cite{openai2024sora,runway,kling} and open-source \cite{yang2024cogvideox,polyak2025moviegen,ma2025stepvideot2v,kong2024hunyuanvideo,wan2.1} T2V models. CogVideoX \cite{yang2024cogvideox} adopts MMDiT to T2V and achieves impressive results in modeling coherent long-duration videos with natural movements. Ma \etal \cite{ma2025stepvideot2v} and Polyak \etal \cite{polyak2025moviegen} scaled the T2V foundation model to $30B$ and demonstrated promising improvements in simulating natural motions. Specifically, Ma \etal \cite{ma2025stepvideot2v} employed a video-based direct preference optimization (DPO) approach \cite{rafailov2024dpo}, namely Video-DPO, to improve the visual quality of generated videos. The recently released open-source models HunyuanVideo \cite{kong2024hunyuanvideo} and Wan2.1 \cite{wan2.1} exhibit much improved video quality and prompt controllability, significantly facilitating the research of video generation in the community. 

\textbf{Downstream Tasks of Visual Generation Models}.
With the advancement in pre-trained T2I and T2V foundation models, researchers have developed various techniques to adapt them to various content creation and manipulation tasks, such as controllable generation \cite{zhang2023controlnet,wu2023tuneavideo}, personalized generation \cite{ruiz2023dreambooth}, editing \cite{brooks2022instructpix2pix,liew2023magicedit}, super-resolution \cite{yang2023pasd}, among others. Zhang \etal \cite{zhang2023controlnet} introduced ControlNet to facilitate various conditional inputs, which, however, requires multiple control modules for different conditions. UniControl \cite{qin2023unicontrol} and UNIC-Adapter \cite{duan2024unic-adapter} enable unified conditional image generation using a single model. InstructPix2Pix \cite{brooks2022instructpix2pix} and MagicBrush \cite{zhang2023magicbrush} offer general-purpose image editing solutions. However, for video tasks, most approaches \cite{wu2023tuneavideo,liew2023magicedit} still follow a single-model single-task framework due to the complexities of video generation. Very recently, Jiang \etal \cite{jiang2025vace} proposed a so-called all-in-one model for multiple visual creation and editing tasks based on pre-trained T2V models \cite{kong2024hunyuanvideo,wan2.1}. Although achieving impressive results, this model is built on pre-train T2V models and treats the other tasks as downstream applications. In contrast, in this work, we train a single model from scratch, which can, however, perform multiple visual generation and manipulation tasks, by effectively utilizing the available training data from different tasks.

\section{Many-for-Many Unified Training}
\label{sec:methd}
\begin{figure}[t!]
\includegraphics[width=0.96\textwidth]{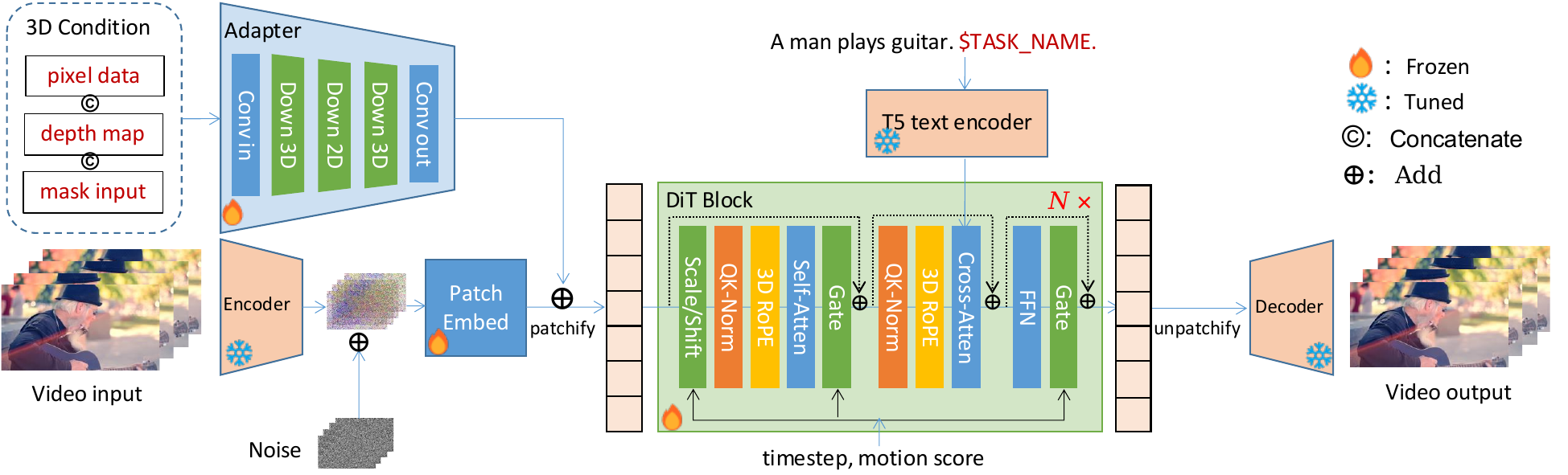}
\caption{Architecture of the proposed Many-for-Many (MfM) unified training framework.}
\centering
\label{fig:arch}
\vspace{-3mm}
\end{figure}

The overall architecture of our Many-for-Many (MfM) unified training framework is illustrated in Figure~\ref{fig:arch}. Basically, our model is a Diffusion Transformer (DiT) \cite{peebles2022dit} with 3D full attention, trained by using the Flow Matching technique \cite{lipman2023fm}. Videos and text prompts are encoded using a video VAE \cite{yang2024cogvideox} and an LLM text encoder \cite{raffel2020t5}, respectively. To mitigate the high reliance on costly annotation of T2V training data and make the best use of existing training data from various visual generation and manipulation tasks, we introduce an effective and lightweight adapter that unifies the various conditions across different tasks. A progressive and joint training strategy is then developed to train a unified model for multiple visual generation and manipulation tasks.  

To accommodate varying computational demands and performance requirements, we design two versions of our model with different sizes, as summarized in Table~\ref{tab:version}. The larger model contains $8$ billion parameters, comprising $40$ layers, $48$ attention heads, and has a model dimension of $3,072$, while the smaller model contains $2$ billion parameters, comprising $28$ layers, $28$ attention heads, and has a model dimension of $1,792$.

\begin{table}[t!]
\centering
\caption{Hyper-parameters of our 2B and 8B model variants.}
\vspace*{1mm}
\normalsize{
\resizebox{0.9\textwidth}{!}{\begin{tabular}{c|c|c|c|c|c}
Model Size & Layers & Attention Heads & Head Dim & FFN Dim & Cross-Attn Dim \\
\hline
2B & 28 & 28 & 64 & 7168 & (1792, 2048) \\
8B & 40 & 48 & 64 & 12288 &  (3072, 2048) \\
\hline
\end{tabular}}}
\label{tab:version}
\vspace{-3mm}
\end{table}

\subsection{Adapter for Different Inputs}
The required model inputs vary significantly across different image and video tasks \cite{openai2024sora,hu2023animateanyone,wu2023tuneavideo,jiang2025vace}. We categorize these inputs based on their dimensions: 0D conditions (\eg, timestep and motion score), 1D conditions (\eg, text), 2D conditions (\eg, image and mask) and 3D conditions (\eg, video and video depths). 0D and 1D conditions are commonly used in DiT, which are embedded using AdaLN and a text encoder, respectively. 2D conditions can be padded to 3D and thus merged into 3D conditions. The 3D conditions include both pixel data (\eg, image and video) and masks, which vary across generation and manipulation (including enhancement) tasks:
\begin{itemize}[leftmargin=*]
    \item \textbf{Generation Tasks}: These tasks require at least one frame to be generated without any frame-wise condition. Examples include T2I (text-to-image), T2V (text-to-video), I2V (image-to-video), video extension, FLF2V (first-last-frame-to-video) and FLC2V (first-last-clip-to-video).
    \item \textbf{Manipulation Tasks}: These tasks require frame-wise conditions. Examples include image/video inpainting/outpainting, image/video colorization, image/video style transfer, single image super-resolution (SISR), video super-resolution (VSR), \etc.
\end{itemize}

\begin{figure}[t!]
\includegraphics[width=0.95\textwidth]{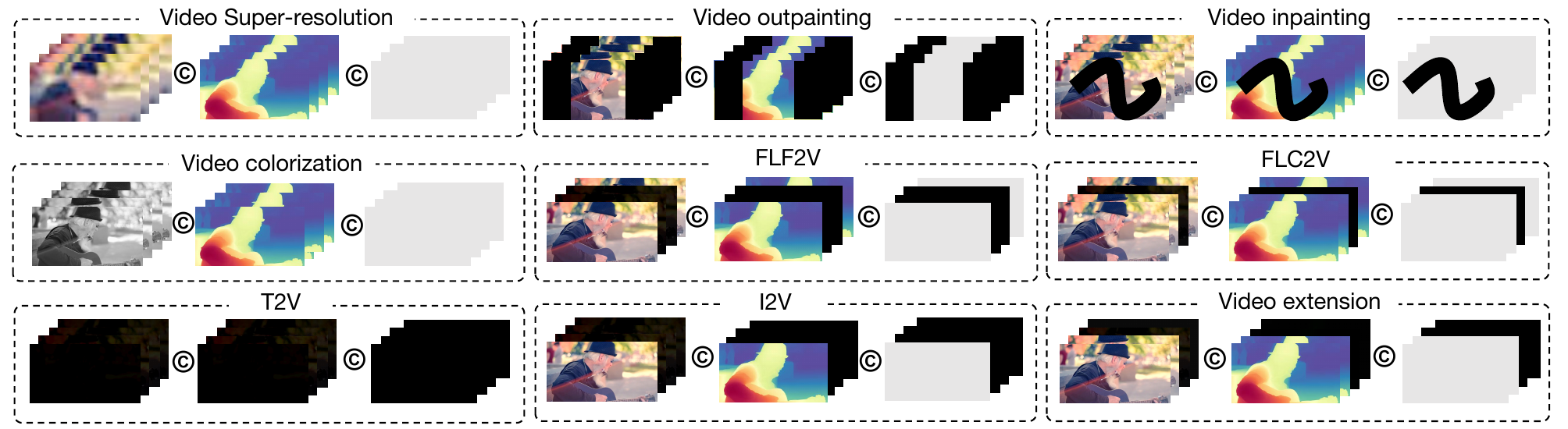}
\vspace{-1mm}
\caption{Example conditional inputs for some visual generation and manipulation tasks. In each task block, from left to right, the conditions are respectively task-oriented pixel data, depth maps, and mask inputs. The mask is composed of binary black and white pixels, with white pixels indicating the regions that are conditioned on pixel data, and black pixels indicating the regions to be generated.}
\centering
\label{fig:tasks}
\vspace{-3mm}
\end{figure}

As illustrated in the upper-left corner of Figure~\ref{fig:arch}, fortunately, we can represent the various inputs in a unified manner, concatenating the pixel, depth map and mask conditions. The depth maps are introduced as a condition to enhance our model's understanding of 3D space. Note that we append the task name (\eg, ``text-to-video'', ``image-to-video'', \etc) to the text prompts to clarify tasks because some of them share a common video mask input, such as VSR and video colorization. Figure~\ref{fig:tasks} illustrates some example inputs for different generation and manipulation tasks. For instance, for the T2V task, the pixel data, depth map, and mask inputs are all set to $0$ so that the task is driven by merely the text prompt. For the task of I2V, only the conditions of the first frame are provided.

Existing visual generation methods typically process the pixel and mask conditions separately — pixel conditions are processed by video VAE, while mask conditions are directly reshaped and interpolated \cite{jiang2025vace,wan2.1}. Although these methods achieve impressive results, they are complex and cannot be easily extended to other types of conditions such as depth maps.
Our proposed adapter unifies all 3D inputs, regardless of their content (\eg, pixel, mask, depth). The adapter comprises several convolution layers and downsampling blocks to adjust the temporal and spatial resolutions. Given a 3D condition input in pixel space $Y \in \mathbb{R}^{T \times H \times W \times C}$, where $\{T, H, W, C\}$ represents the frame number, height, width, and channel number, the adapter converts it into a feature map $y \in \mathbb{R}^{t \times h \times w \times c}$, which  shares the same spatial and temporal resolution as the latent space of the video VAE and is simply added to the latent video feature. Given the video VAE's $8\times8$ spatial and $4\times$ temporal compression ratios \cite{yang2024cogvideox}, we have $t = T/4, h = H/8, w = W/8$. The proposed architecture can be easily adjusted according to the compression ratios of alternative video VAEs. 

\subsection{Transformer with 3D Full Attention}
\textbf{3D Full Attention}.
Early video generation approaches \cite{blattmann2023svd,guo2023animatediff} are typically built on pre-trained T2I models, which often use separate spatial and temporal attention to reduce computational complexity. However, such methods are suboptimal for modeling natural motions, as demonstrated by large-scale experiments \cite{yang2024cogvideox}. In recent works \cite{openai2024sora,yang2024cogvideox,ma2025stepvideot2v,kong2024hunyuanvideo}, 3D full attention has become widely adopted and shown superiority in generating videos with smooth and consistent motions. In this work, we incorporate the transformer block design from GoKu \cite{chen2025goku}, which consists of a self-attention module to capture relationships within input sequences, a cross-attention layer to include text embeddings, and an adaptive layer normalization (AdaLN) operation to embed timestep and motion score information.

\textbf{3D RoPE}.
Rotary Position Embedding (RoPE) \cite{su2021rope} encodes positional information and enables the model to understand both the absolute position of tokens and their relative distances, demonstrating powerful ability in capturing inter-token relationships, particularly for long sequences in LLMs. We extend it to 3D RoPE by applying 1D RoPE to each temporal ($t$) and spatial ($h,w$) dimension, then concatenating the encodings. Specifically, for 3D video data ($t,h,w$), each dimension occupies 
$2/8$, $3/8$, and $3/8$ of the hidden state channels, respectively. We apply 3D RoPE for both image and video tokens. Due to the exceptional extrapolation capabilities of RoPE, the proposed 3D RoPE can effectively handle videos with varying resolutions and lengths.

\textbf{Q-K Normalization}.
Previous methods \cite{esser2024sd3,dehghani2023vit} have shown that the training of large transformer models can encounter numerical instability due to the uncontrollable growth in attention entropy. To address this issue, following \cite{esser2024sd3,dehghani2023vit}, we adopt RMSNorm \cite{zhang2019rmsnorm} and implement Query-Key Normalization (QK-Norm) to stabilize the training process.

\subsection{Training Details}
\label{sec:train}
\textbf{Flow Matching}.
During model training, we employ Rectified Flow (RF) to optimize the network due to its superior performance \cite{lipman2023fm,esser2024sd3,chen2025goku}. In each training step, a video input $X_0$, Gaussian noise $\epsilon \sim \mathcal{N}(0,1)$, and a timestep $t \in [0,1]$ are randomly sampled. The model input $X_t$ is calculated as a linear interpolation between $\epsilon$ and $X_0$: \begin{equation} 
X_t = (1-t)X_0 + t\epsilon. 
\end{equation}
The model is trained to approximate the ground-truth velocity $V_t = \frac{dX_t}{dt} = \epsilon - X_0$, which represents the change rate of $X_t$ with respect to timestep $t$, capturing the change direction and magnitude from $\epsilon$ to $X_0$. Given conditions of motion score $ms$, text prompt $c$, and 3D conditional input $Y$, we train our model $\mu_\theta$ to predict the velocity $V_t$. The optimization objective $\mathcal{L}$ is defined as: 
\begin{equation}
\mathcal{L} = \mathbb{E}_{t,X_0,\epsilon \sim \mathcal{N}(0,1),ms,c,Y} |\mu_\theta(t,X_t,ms,c,Y) - V_t|^2. 
\end{equation}
Following SD3 \cite{esser2024sd3}, we use Logit-Normal Sampling in training.

\textbf{Multi-Task Joint Learning}. While our model is primarily designed for video generation, we leverage a large volume of image data into the training process. Following the strategy of existing T2V foundation models \cite{openai2024sora,kong2024hunyuanvideo,chen2025goku}, we progressively adjust the image-to-video ratio throughout the training. Initially, we train with pure text-image pairs to establish a connection between textual prompts and high-level visual semantics. As training progresses, we inject video data, gradually decreasing the image-to-video ratio to 0.1. This image-video joint learning strategy expands our training data and enables our model to tackle various image tasks, including T2I and SISR.

Unlike standard T2V foundation models, our training data include a substantial portion of low-resolution, watermarked, text-dominated, and concisely captioned data. 
To effectively utilize the available training data, we implement multi-task learning, thanks to our proposed conditional adapter. At each training step, we randomly sample a video input, assign a set of qualified tasks that fits it, and select one task to construct the conditional input for training. For each qualified task set, the selection probability of tasks like T2I, T2V, and I2V is tripled compared to other tasks, ensuring that the learning process pay more attention to more challenging problems.

\textbf{Resolution Progressive Training}.
Our training pipeline, equipped with the image-video and multi-task joint learning strategy, is structured into multiple stages with progressively increased spatial and temporal resolutions. Initially, we train our model on low-resolution data (\eg, $49\times128\times224$) at a low computational cost. We then increase the resolution to $89\times352 \times640$ to enhance the model’s fine-grained understanding of text-motion relationships. Subsequently, the training resolution is increased to $97\times720\times1280$ to capture intricate details. Finally, we conclude the training pipeline with a multi-resolution stage using NaViT \cite{dehghani2023navit}. In this stage, the model is fed high-quality videos with their native aspect ratios, dynamically adjusting the durations to limit the total sequence length. This multi-resolution fine-tuning stage enables our model to generate videos at arbitrary resolutions. During training, we randomly replace 10\% (30\%) of text prompts with null-text prompts for T2V\//I. For tasks other than T2V\//I, we randomly zero the 3D conditional inputs with a chance of 10\%.

\begin{table}[t!]
\centering
\caption{Resolution progressive training recipe for 8B MfM.}
\vspace*{1mm}
\normalsize{
\resizebox{0.95\textwidth}{!}{\begin{tabular}{c|c|c|c|c|c|c}
Training Stage & Dataset & SP & bs/GPU & Learning rate & \#iters & \#seen samples \\
\hline
128px & \makecell{160M images \\ 120M videos} & 1 & 16 & 1e-4 & 170k & 700M \\
\hline
360px & \makecell{160M images \\ 120M videos} & 1 & 2 & 8e-5 & 100k & 100M \\
\hline
720px & \makecell{160M images \\ 10M videos} & 2 & 1 & 5e-5 & 50k & 12M \\
\hline
Multi-res & \makecell{160M images \\ 5M videos} & 2 & 1 & 5e-5 & 40k & 5M \\
\hline
\end{tabular}}}
\label{tab:recipe}
\vspace{-5mm}
\end{table}

\section{Experiments}
\label{sec:exp}
\subsection{Experiment Setup}
As described in Section~\ref{sec:train}, we employ a progressive pipeline to improve training efficiency and model scalability. The detailed training recipe for our 8B model is summarized in Table~\ref{tab:recipe}. We adopt Fully Sharded Data Parallelism (FSDP) \cite{zhao2023fsdp} and an optional Sequence-Parallelism (SP) \cite{li2022sp,jacobs2023sp} to achieve efficient and scalable training of MfM. 

\textbf{Training Data Preparation}. 
Our training data are collected from a variety of sources, including publicly available academic datasets, Internet resources, and proprietary datasets. We adopt a data curation pipeline similar to GoKu \cite{chen2025goku} to filter the collected data, obtaining $160$M HQ text-image pairs and $40$M HQ text-video pairs. We also retrieve $80$M relatively LQ text-video pairs for training. 
We utilize RAFT \cite{teed2020raft} to obtain motion scores by computing the mean optical flow of video clips, which are integrated into our MfM model training via AdaLN.

Note that we use significantly fewer text-video pairs than the main T2V models \cite{polyak2025moviegen,ma2025stepvideot2v,kong2024hunyuanvideo,wan2.1} to train our MfM model. However, our MfM framework leverages a multi-task data augmentation strategy to expand the effective training data distribution. Specifically, for video inpainting (VINP), we apply random masks to interior regions; for video outpainting (VOUTP), we generate boundary masks; for video colorization (VCOLOR), we convert the original videos to grayscale; for video extension (VEXT), we extract the first 8-16 frames as conditioning input; for first-last-frame-to-video (FLF2V), we sample the first and last frames as conditioning signals; for first-last-clip-to-video (FLC2V), we sample both initial 8-16 frames and final 8-16 frames as conditioning signals; for video super-resolution (VSR), we apply random downsampling factors between 2-6×; and for video editing (VEDIT), we incorporate style transfer pairs from the InsViE-1M dataset \cite{wu2025insvie}. Similar pre-processing is applied to image data, creating image-based training data. Through such a systematic data augmentation strategy, we significantly expand the model's exposure to diverse conditioning scenarios without requiring additional data collection.

For all tasks, we employ a lightweight depth model \cite{yang2024depth_anything_v2} to predict the depth maps of the inputs on the fly. We concatenate these depth maps into the 3D conditional inputs as depicted in Figure~\ref{fig:tasks}. 

\textbf{Evaluation}.
To evaluate MfM's performance on the fundamental T2V and I2V tasks, we utilize the widely used VBench \cite{huang2024vbench}, which provides standardized assessment on video quality across multiple dimensions. For evaluation on the model's multi-task capacity, while a VACE-Benchmark is mentioned in \cite{jiang2025vace}, only one video is open-sourced for each task, and many tasks supported by MfM are not involved in VACE. Therefore, we build an MfM-benchmark, which comprises 480 samples (30 per task) distributed across 16 distinct generation/manipulation tasks. The details of MfM-benchmark can be found in the \textbf{Appendix}. For all experiments, we maintain the same MfM inference parameters: 50 diffusion steps with a classifier-free guidance scale of 9.0.

Regarding evaluation metrics, on VBench we adopt a comprehensive set of perceptual metrics: aesthetic quality, imaging quality, motion smoothness, dynamic degree, object class accuracy, multiple object handling, color fidelity, spatial relationship preservation, scene consistency, appearance style, temporal style, and overall consistency (higher scores indicate better performance across all metrics). Meanwhile, we rank the competitors for each metric and calculate the average rank over all metrics for each method. For some tasks on MfM-benchmark, we also use reference-based metrics, including FID, PSNR, SSIM, and LPIPS, to quantify the fidelity of generated content.


\subsection{Experimental Results on T2V and I2V}

\begin{table}[!t]
\vspace{-4mm}
\centering
\caption{Quantitative comparison of T2V generation performance on the VBench-T2V benchmark. Comparison baselines are selected from VBench leaderboard. For each dimension, the best result is in bold, the second best result is underscored and the third best result is italic. (Aesth: Aesthetic Quality; Img: Imaging Quality; Mul.Obj: Multiple Objects; Temp: Temporal Style; Consist: Overall Consistency; Avg: Average Ranking.)}
\small  
\setlength{\tabcolsep}{2.0pt}  
\normalsize{
\resizebox{\textwidth}{!}{
\begin{tabular}{l|*{12}{c}|c}
\toprule
\textbf{Model} & 
\textbf{\shortstack{Motion}} & 
\textbf{\shortstack{Dynamic}} & 
\textbf{\shortstack{Aesth.}} & 
\textbf{\shortstack{Img.}} & 
\textbf{\shortstack{Object}} & 
\textbf{\shortstack{Mul.Obj.}} & 
\textbf{Color} & 
\textbf{\shortstack{Spatial}} & 
\textbf{Scene} & 
\textbf{\shortstack{Appear.}} & 
\textbf{\shortstack{Temp.}} & 
\textbf{\shortstack{Consist.}} & 
\textbf{Avg.} \\
\midrule
MfM & 0.983 & \underline{0.819} & \underline{0.645} & 0.662 & \textit{0.927} & \underline{0.782} & 0.836 & \underline{0.802} & \underline{0.546} & \textbf{0.251} & \textit{0.247} & \textbf{0.277} & \textbf{3.04} \\
Wan2.1~\cite{wan2.1}    & 0.969 & \textbf{0.943} & 0.615 & \textit{0.672} & \textbf{0.942} & \textbf{0.814} & 0.877 & \textbf{0.810} & 0.536 & 0.211 & \textbf{0.256} & \underline{0.274} & \textit{3.79} \\
Hunyuan~\cite{kong2024hunyuanvideo}  & \textit{0.989} & 0.708 & 0.603 & \underline{0.675} & 0.861 & 0.685 & \textbf{0.916} & 0.686 & 0.538 & 0.198 & 0.238 & 0.264 & 5.33 \\
Sora~\cite{openai2024sora}  & 0.987 & \textit{0.799} & \textit{0.634} & \textbf{0.682} & \underline{0.939} & \textit{0.708} & 0.801 & 0.742 & \textbf{0.569} & \underline{0.247} & 0.250 & 0.262 & \underline{3.67} \\
Gen-3~\cite{runway}  & \underline{0.992} & 0.601 & 0.633 & 0.668 & 0.878 & 0.536 & 0.809 & 0.650 & \textit{0.545} & 0.243 & \textit{0.247} & \textit{0.266} & 4.96 \\
PikaLabs~\cite{pika}  & \textbf{0.995} & 0.475 & 0.620 & 0.618 & 0.887 & 0.430 & \underline{0.905} & 0.610 & 0.498 & 0.222 & 0.242 & 0.259 & 6.54 \\
LTX-Video~\cite{hacohen2024ltxvideo}  & \textit{0.989} & 0.543 & 0.598 & 0.602 & 0.834 & 0.454 & 0.814 & 0.654 & 0.510 & 0.214 & 0.226 & 0.251 & 7.67 \\
CogVideoX1.5~\cite{yang2024cogvideox}  & 0.981 & 0.561 & 0.620 & 0.653 & 0.834 & 0.672 & \textit{0.884} & \textit{0.794} & 0.532 & \textit{0.246} & \underline{0.254} & \underline{0.274} & 5.04 \\
EasyAnimate~\cite{xu2024easyanimate}  & 0.980 & 0.571 & \textbf{0.694} & 0.585 & 0.895 & 0.668 & 0.778 & 0.761 & 0.543 & 0.230 & 0.246 & 0.264 & 4.96 \\
\bottomrule
\end{tabular}}}
\vspace{-2mm}
\label{tab:t2v}
\end{table}

\begin{table}[tbp]
\vspace{-2mm}
\centering
\caption{Quantitative comparison of I2V generation performance on the VBench-I2V benchmark. Comparison baselines are selected from VBench leaderboard. For each dimension, the best result is in bold and the second best result is underscored. (IS. Consist: Image Subject Consistency; IB. Consist: Image Background Consistency.)}
\small
\setlength{\tabcolsep}{4pt}  
\begin{tabular}{l|*{6}{c}|c}
\toprule
\textbf{Model} & \textbf{IS. Consist.} & \textbf{IB. Consist.} & \textbf{Motion} & \textbf{Dynamic} & \textbf{Aesth.} & \textbf{Img.} & \textbf{Avg.} \\

\midrule
MfM & \textit{0.982} & \underline{0.991} & \textit{0.987} & \textit{0.613} & 0.608 & \textbf{0.718} & \textbf{3.33} \\
Wanx-I2V~\cite{wan2.1} & 0.973 & 0.981 & 0.978 & \underline{0.678} & 0.615 & 0.708 & 5.50 \\
Hunyuan-I2V~\cite{kong2024hunyuanvideo} & \textbf{0.988} & \textbf{0.992} & \textbf{0.994} & 0.239 & 0.617 & 0.700 & \textit{3.67} \\
Magi-1~\cite{magi1}  & \underline{0.983} & 0.990 & 0.986 & \textbf{0.682} & \textit{0.647} & 0.697 & \underline{3.50} \\
Step-Video~\cite{ma2025stepvideot2v} & 0.978 & 0.986 & \underline{0.992} & 0.487 & 0.622 & 0.704 & 3.83 \\
DynamicCrafter~\cite{xing2023dynamicrafter} & 0.981 & 0.986 & 0.973 & 0.474 & \textbf{0.664} & 0.693 & 5.33 \\
VideoCrafter-I2V~\cite{chen2024videocrafter2} & 0.911 & 0.913 & 0.980 & 0.226 & 0.607 & \underline{0.716} & 7.83 \\
I2VGen-XL~\cite{zhang2023i2vgen} & 0.975 & 0.976 & 0.983 & 0.249 & \underline{0.653} & 0.698 & 5.83 \\
CogvideoX-I2V~\cite{yang2024cogvideox} & 0.971 & 0.967 & 0.984 & 0.331 & 0.618 & 0.700 & 6.17 \\
ConsistI2V~\cite{ren2024consisti2v} & 0.958 & 0.959 & 0.973 & 0.186 & 0.590 & 0.669 & 9.50 \\
\bottomrule
\end{tabular}
\vspace{-5mm}
\label{tab:I2V}
\end{table}

\begin{figure}[t]
\includegraphics[width=0.96\textwidth]{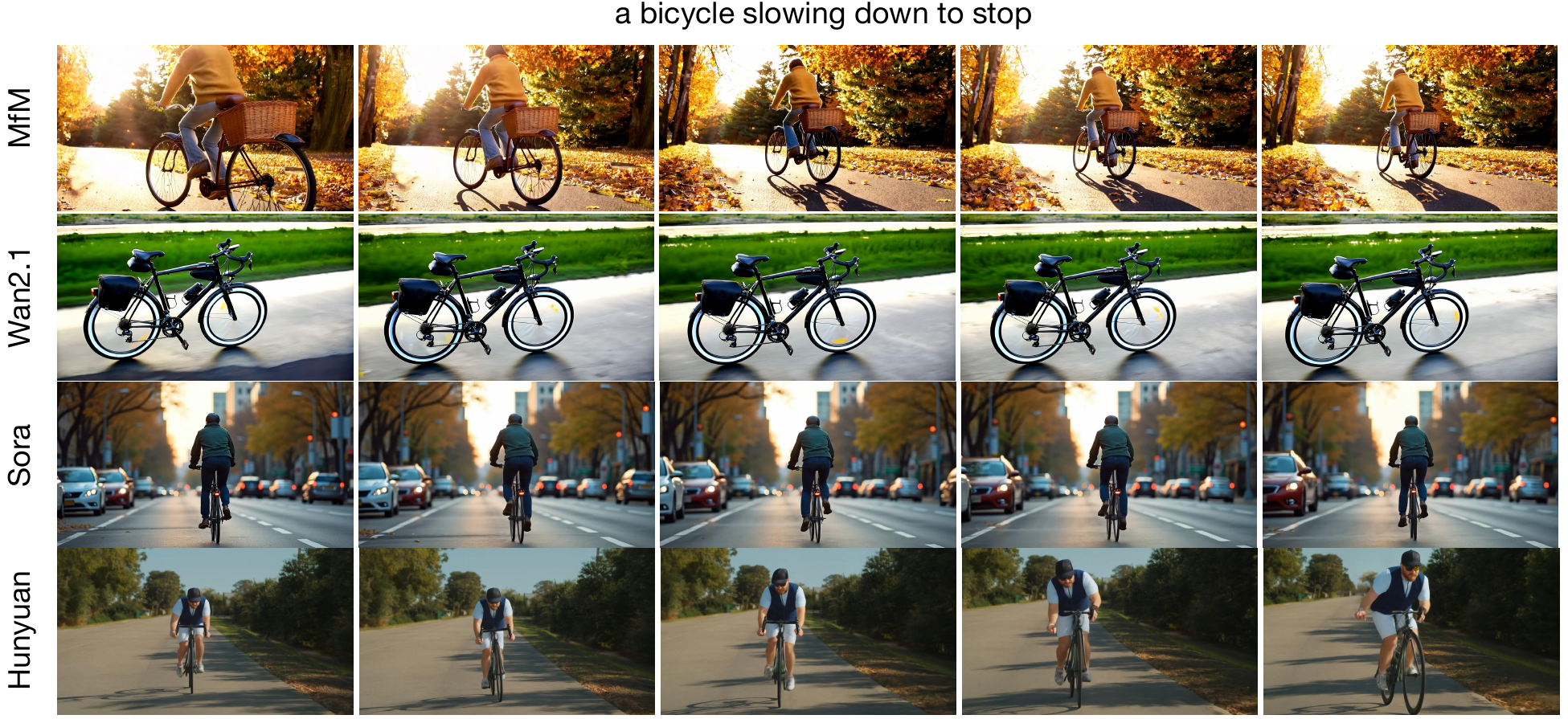}
\caption{Qualitative comparison of T2V generation results on the prompt "a bicycle slowing down to stop." More visual comparisons are provided in the \textbf{Appendix}.}
\centering
\label{fig:visualt2v}
\vspace{-6mm}
\end{figure}

Since most of the existing methods use two separate models for T2V and I2V tasks, we present the quantitative comparison in two tables. The results of T2V are shown in Table~\ref{tab:t2v}.  We can see that MfM achieves the best average rank (3.04) among all evaluated models. In particular, MfM exhibits well-balanced performance across multiple dimensions, ranking the best in appearance and overall consistency, and the second in dynamic degree, aesthetic quality, multiple object generation, and spatial relation generation, which are essential for producing visually coherent videos aligned with textual descriptions. In comparison, the larger models such as Wan2.1 (14B), Hunyuan (13B) and the commercial models such as Sora can achieve impressive scores in specific dimensions, but their overall performance is compromised by notable weaknesses in other dimensions. For instance, Wan2.1 ranks last in motion smoothness, while Hunyuan shows deficiencies in appearance style, resulting in jerky movements, visual distortions, or monotonous video style in some scenarios. 
The visual comparison can be found in Figure~\ref{fig:visualt2v}, where Wan2.1 generates a bicycle without a rider and fails to depict the slowing motion instruction given in the prompt. Similarly, Sora and Hunyuan fail to accurately represent the slowing motion. Hunyuan also exhibits distortion in the bicycle's handlebars as the sequence progresses. Our MfM successfully generates a motion-consistent video with the bicycle correctly slowing down, demonstrating superior temporal coherency.

\begin{table}[tbp]
\centering
\caption{Performance comparison on multiple video manipulation tasks.}
\small
\setlength{\tabcolsep}{3.8pt}
\normalsize{
\resizebox{\textwidth}{!}{
\begin{tabular}{ll|cccc|ccccc}
\toprule
\multirow{2}{*}{\textbf{Task}} & \multirow{2}{*}{\textbf{Method}} & \multicolumn{4}{c|}{\textbf{Reference-based Metrics}} & \multicolumn{5}{c}{\textbf{No-reference Perceptual Metrics}} \\
\cmidrule(lr){3-6} \cmidrule(lr){7-11}
& & \textbf{FID↓} & \textbf{PSNR↑} & \textbf{SSIM↑} & \textbf{LPIPS↓} & \textbf{Aesth.} & \textbf{Img.} & \textbf{Motion} & \textbf{Consist.} & \textbf{Temp.} \\
\midrule
\multirow{3}{*}{VINP} & MfM & \textbf{51.03} & \textbf{21.31} & 0.830 & \textbf{0.112} & 0.560 & \textbf{0.767} & \textbf{0.994} & 0.220 & \textbf{0.985} \\
& VACE\cite{jiang2025vace} & 67.51 & 17.41 & 0.534 & 0.256 & \textbf{0.569} & 0.757 & 0.990 & \textbf{0.224} & 0.983 \\
& ProPainter\cite{zhou2023propainter} & 119.93 & 20.40 & \textbf{0.880} & 0.118 & 0.417 & 0.739 & 0.992 & 0.206 & \textbf{0.985} \\
\midrule
\multirow{4}{*}{VOUTP} & MfM & \textbf{44.15} & \textbf{18.21} & \textbf{0.733} & \textbf{0.168} & 0.539 & \textbf{0.745} & \textbf{0.992} & \textbf{0.216} & \textbf{0.974} \\
& VACE\cite{jiang2025vace} & 54.34 & 16.16 & 0.500 & 0.310 & \textbf{0.567} & 0.736 & 0.987 & 0.211 & 0.971 \\
& FYC\cite{chen2024follow} & 94.69 & 14.49 & 0.416 & 0.414 & 0.550 & 0.736 & 0.988 & 0.211 & 0.971 \\
& M3DDM\cite{fan2023M3DDm} & 174.61 & 17.96 & 0.571 & 0.475 & 0.484 & 0.671 & 0.982 & 0.214 & 0.972 \\
\midrule
\multirow{3}{*}{FLF2V} & MfM & \textbf{31.98} & \textbf{19.95} & \textbf{0.583} & \textbf{0.203} & \textbf{0.525} & 0.730 & 0.981 & 0.225 & 0.966 \\
& Wanx\cite{wan2.1} & 38.24 & 18.28 & 0.512 & 0.244 & 0.520 & \textbf{0.742} & \textbf{0.990} & \textbf{0.229} & \textbf{0.978} \\
& Hunyuan\cite{kong2024hunyuanvideo} & 118.18 & 10.17 & 0.372 & 0.419 & 0.476 & 0.598 & 0.992 & 0.225 & 0.985 \\
\midrule
\multirow{3}{*}{VCOLOR} & MfM & \textbf{76.54} & 17.93 & 0.810 & 0.176 & 0.582 & 0.756 & \textbf{0.993} & \textbf{0.230} & \textbf{0.985} \\
& colormnet\cite{yang2024colormnet} & 77.08 & 17.47 & \textbf{0.812} & \textbf{0.160} & \textbf{0.594} & \textbf{0.758} & 0.990 & \textbf{0.230} & 0.980 \\
& TCVC\cite{zhang2023TCVC} & 82.42 & \textbf{20.69} & 0.699 & 0.201 & 0.553 & 0.720 & 0.991 & 0.228 & 0.984 \\
\bottomrule
\end{tabular}}}

\vspace{-4mm}
\label{tab:multitask}
\end{table}

The results of I2V are shown in Table~\ref{tab:I2V}. We see that MfM also achieves the best average rank (3.33). In particular, it excels in imaging quality and achieves very balanced performance across static consistency and dynamic generation. In comparison, although Hunyuan-I2V achieves the highest scores in consistency and motion smoothness, its performance in dynamic degree and aesthetic qualities is substantially lower, resulting in an average rank of only 3.67, lower than MfM and Magi-1. Visual comparisons of I2V generation are provided in the \textbf{Appendix}.

Finally, it is worth mentioning that our MfM achieves competitive results in both T2V and I2V generation tasks using a \textbf{single unified model}, while previous approaches such as Wan and Hunyuan rely on separate specialized models for each generation paradigm. The unified nature of MfM reduces overall model parameters and ensures consistent visual quality between text and image conditioning.

\subsection{Performance on Multiple Video Manipulation Tasks}

Beyond T2V and I2V generation, our MfM supports 16 distinct tasks through a unified model, including 10 video-related tasks (T2V, I2V, VINP, VOUTP, FLF2V, VCOLOR, FLC2V, VEXT, VSR and VEDIT) and 6 image-related tasks (T2I, image inpainting, image outpainting, style transfer, image colorization, and image super-resolution). Given that our primary focus is on video generation and manipulation, while the image tasks and data are used to aid video task training, we perform evaluation on a subset of video tasks with established baselines for comparison. Specifically, we select four representative tasks, including video inpainting, video outpainting, video transition and video colorization, for experiment since they have competitive baseline models and standardized evaluation protocols. Table~\ref{tab:multitask} presents quantitative comparisons on our established MfM-Benchmark. Visual comparisons can be found in the \textbf{Appendix}.

Our experimental results demonstrate MfM's excellent versatility and effectiveness as a unified video foundation model across diverse manipulation tasks. First, MfM shows consistent advantages in reference-based metrics. In particular, it achieves FID improvements ranging from 16.4\% to 72.9\% over the specialized models of these tasks. In addition to reference-based metrics, MfM exhibits impressive temporal coherence, which demonstrates MfM's strong ability to seamlessly transition between different operations: inferring complex motion change from two frames, preserving spatial coherence during region manipulation, and maintaining consistent appearance while modifying visual attributes. 
Meanwhile, with MfM the knowledge learned from one task can benefit another task. For example, the capability developed for handling boundaries in outpainting can enhance performance in inpainting; similarly, the motion-inference ability required for video translation contributes to the temporal coherence observed in colorization tasks.
In summary, MfM can effectively capture the principles underlying diverse video manipulation tasks and achieve competitive performance without requiring separate architectures for each manipulation paradigm. 

\subsection{The Benefit of Multi-task Training to Video Generation} 

\begin{table}[tbp]
\centering
\caption{Ablation study on multi-task training versus single-task training on VBench-T2V.}
\small  
\setlength{\tabcolsep}{3.5pt}  
\normalsize{
\resizebox{\textwidth}{!}{
\begin{tabular}{l|*{12}{c}}
\toprule
\textbf{Model} & 
\textbf{\shortstack{Motion}} & 
\textbf{\shortstack{Dynamic}} & 
\textbf{\shortstack{Aesth.}} & 
\textbf{\shortstack{Img.}} & 
\textbf{\shortstack{Object}} & 
\textbf{\shortstack{Mul.Obj.}} & 
\textbf{Color} & 
\textbf{\shortstack{Spatial}} & 
\textbf{Scene} & 
\textbf{\shortstack{Appear.}} & 
\textbf{\shortstack{Temp.}} & 
\textbf{\shortstack{Consist.}} \\
\midrule
T2V w/ MfM & 0.983 & \textbf{0.819} & \textbf{0.645} & \textbf{0.662} & 0.927 & 0.782 & 0.836 & 0.802 & \textbf{0.546} & \textbf{0.251} & \textbf{0.247} & \textbf{0.277}\\
T2V w/o MfM  & \textbf{0.989} & 0.402 & 0.628 & 0.641 & \textbf{0.935} & \textbf{0.887} & \textbf{0.873} & \textbf{0.882} & 0.534 & 0.250 & 0.221 & 0.272 \\

\bottomrule
\end{tabular}}}
\vspace{-4mm}
\label{tab:ablation}
\end{table}

To validate that our multi-task training strategy can benefit video generation tasks when datasets are limited, we conduct an ablation study by training a T2V model with and without our MfM strategy. For the T2V model without MfM, the training data contain 160M text-image pairs and 120M text-video pairs. The results on the VBench-T2V benchmark are shown in Table~\ref{tab:ablation}.
We can see that the T2V models with (w/) and without (w/o) MfM achieve similar scores in most dimensions, but there is a striking difference in the dynamic degree, where T2V w/ MfM scores 0.819 and T2V w/o MfM scores only 0.402. This substantial difference suggests that exposure to diverse video manipulation tasks significantly improves the model's ability to generate dynamic content with appropriate motion complexity. Visual illustrations are presented in the \textbf{Appendix}. Meanwhile, T2V w/ MfM demonstrates more balanced performance across the evaluation spectrum, particularly in quality-related metrics (aesthetic quality, imaging quality). In summary, by learning from diverse conditioning types and generation tasks, the MfM model develops a more comprehensive understanding of video dynamics while maintaining excellent video imaging quality.

\section{Conclusion}

In this work, we introduced MfM (Many-for-Many), a unified video foundation model capable of handling diverse visual generation and manipulation tasks through a single parameter-efficient architecture. Specifically, we designed a lightweight adapter to effectively unify various 2D and 3D conditions into a uniform representational space, enabling seamless integration into our video generation pipeline. By employing progressive joint image-video learning and multi-task training strategies, we not only enabled multiple visual generation and manipulation capabilities within a single model but also transferred the knowledge from other image tasks to video generation. This knowledge sharing significantly reduced the required amount of costly text-to-video training data and enhanced the fundamental video generation capabilities. As validated in our experiments, MfM achieved competitive or superior performance compared to specialized models and even commercial systems while using much fewer training data and model parameters.

\textbf{Limitations}. Despite the demonstrated effectiveness, we acknowledge certain limitations of our proposed MfM. Currently, MfM processes 1D conditions (text) and 2D/3D conditions (masks, pixels, depth) separately before implicitly fusing them through self-attention in DiT blocks. In future work, we will explore the use of vision-language models rather than text-only encoders to perform explicit multimodal fusion earlier in the pipeline, which could enhance performance on tasks requiring comprehensive understanding of complex input conditions.

\small
\bibliographystyle{ieeenat_fullname}
\bibliography{reference}
\clearpage

\clearpage
\appendix

\renewcommand{\thesection}{} 
\begin{center}
{\Large\bf Technical Appendices and Supplementary Material}
\end{center}
\renewcommand{\thesection}{\Alph{section}} 




In this appendix, we provide visual demonstrations and the following supporting materials to the main paper:

\begin{itemize}[leftmargin=1em]
    \item The details of MfM-Benchmark construction (referring to Sec. 4.1 in the main paper);
    \item Visual results of T2V generation on VBench (referring to Sec. 4.2 in the main paper);
    \item Visual results of I2V generation on VBench (referring to Sec. 4.2 in the main paper);
    \item User study of I2V generation;
    \item Visual results of multi-task generation on MfM-Benchmark (referring to Sec. 4.3 in the main paper);
    \item Visual results of T2V generation with or without multi-task training (referring to Sec. 4.4 in the main paper);
    \item Failure cases of MfM;
    \item Broader impacts and safeguards.
\end{itemize}

For better viewing experience, we uploaded the video demos to a website \blueContent{\href{https://leeruibin.github.io/MfMPage/}{https://leeruibin.github.io/MfMPage/}}, where the videos can be played directly in the browser.
Note that, due to the significant number of high-quality video files included in our demonstrations, initial page loading may require several minutes to complete. We appreciate your patience during this process, as the complete visual experience is essential to understand the capabilities and performance of our approach.

\section{The details of MfM-Benchmark construction}

First, we collected 1500 videos of 1280×720 of resolution and their accompanying captions from Pexels~\cite{pexels2025}, selecting only those containing more than 97 frames. We then applied a two-stage quality filtering process: (1) removing blurry videos by calculating the CV2.Laplacian~\cite{opencv_library} score for each frame and excluding those below a threshold of 200, and (2) evaluating motion dynamics using RAFT~\cite{teed2020raft} and retaining only videos with motion scores exceeding 5. This filtering resulted in our final dataset of 480 high-quality videos, which serve as ground-truth for reference-based metrics. We standardized each video to 97 frames and divided them into 16 segments for consistent processing.

For task-specific data preparation, we applied the enhancement pipeline described in Section 4.1 of the main paper. Specifically:

\begin{enumerate}

\item Text-to-Video (T2V): We used the original captions as conditioning input.
\item Image-to-Video (I2V): We used the first frame and caption as conditioning input.
\item Video Extension (VEXT): We extracted the first 8 frames as conditioning input to generate the remaining frames.
\item Video Inpainting (VINP): We applied random masks to interior regions covering 1/9 to 1/4 of the total pixels.
\item Video Outpainting (VOUTP): We generated boundary masks covering 1/8 to 1/4 of the total width/height.
\item Video Colorization (VCOLOR): We converted the ground-truth videos to grayscale.
\item First-Last-Frame-to-Video (FLF2V): We used the first and last frames as conditioning input to generate the intermediate 95 frames.
\item First-Last-Clip-to-Video (FLC2V): We used the first 8 frames and last 8 frames as conditioning input.
\item Video Super-Resolution (VSR): We applied random downsampling factors between $2\times$ and $6\times$ and used the downsampled videos as conditioning input.
\item Video Editing (VEDIT): We used the original videos as conditioning input, replacing the original captions with style instruction prompts (\textit{e.g.}, "change the video to oil painting style").
\item Text-to-Image (T2I): We used the first frame at the ground-truth.
\item Image Super-Resolution (SISR): We used the first frames at the ground truth an downscaled them with downsampling factors between $2\times$ and $6\times$.
\item Image Inpainting (IINP): We sampled the first frames and randomly masked them like VINP.
\item Image Outpainting (IOUTP): we sampled the first frames and randomly masked them like VOUTP.
\item Image Coloraization (ICOLOR): We sampled the first frames and converted them to grayscale.
\item Image Editing (IEdit): We sampled the first frames and replaced the original captions with instruction prompts such as VEDIT.
\end{enumerate}

\begin{figure}[t]
\includegraphics[width=0.96\textwidth]{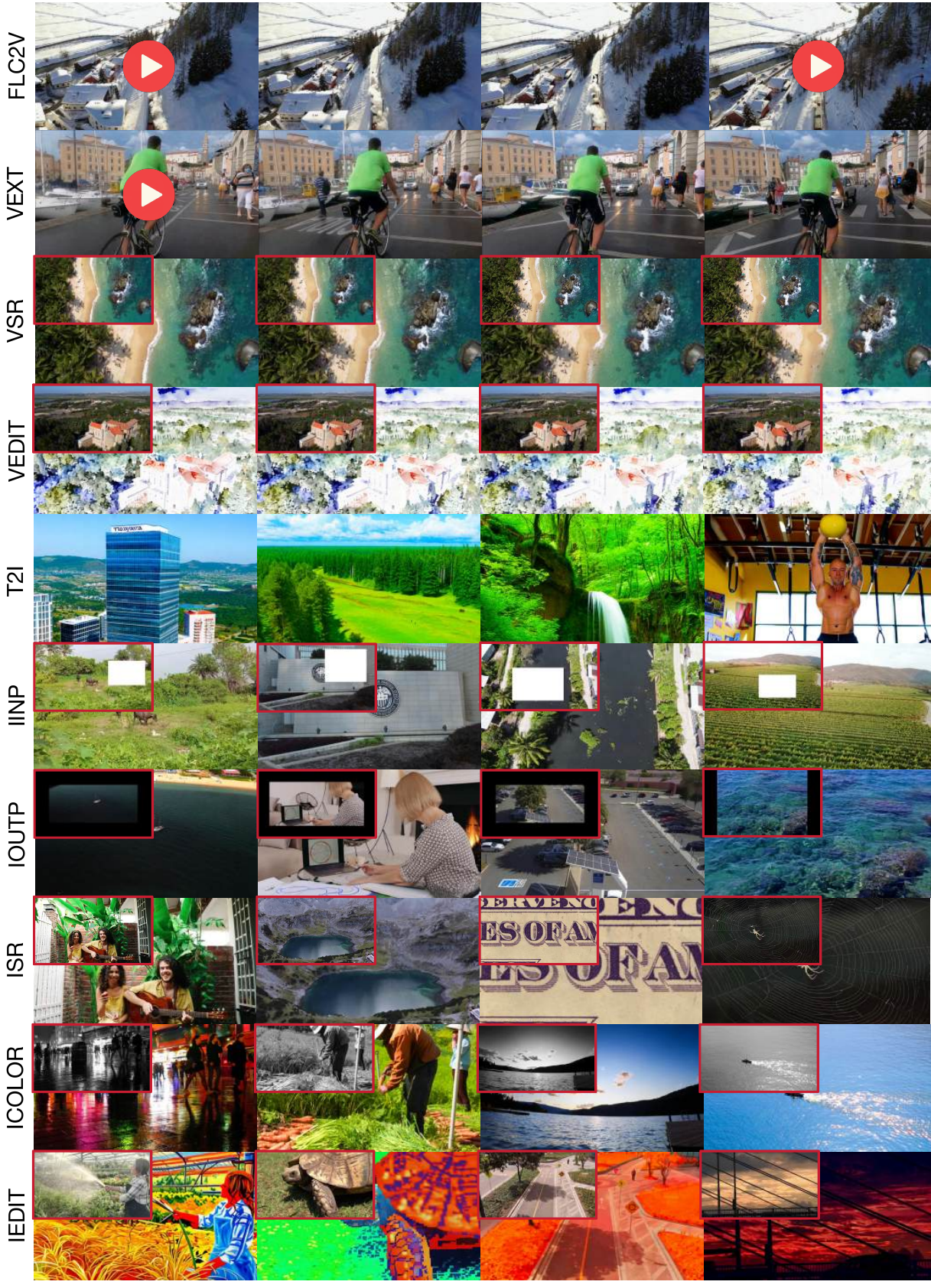}
\caption{Visual illustrations for different tasks supported by our MfM.}
\centering
\label{fig:other}
\vspace{-6mm}
\end{figure}

Visual illustrations of these tasks are shown in Figure~\ref{fig:other}, Figure~\ref{fig:vinp}, Figure~\ref{fig:voup}, Figure~\ref{fig:vcolor}, Figure~\ref{fig:FLF2V}. Video demonstrations are also available at \blueContent{\href{https://leeruibin.github.io/MfMPage/}{https://leeruibin.github.io/MfMPage/}}.

\section{Visual results of T2V generation on VBench}

\begin{figure}[t]
\includegraphics[width=0.96\textwidth]{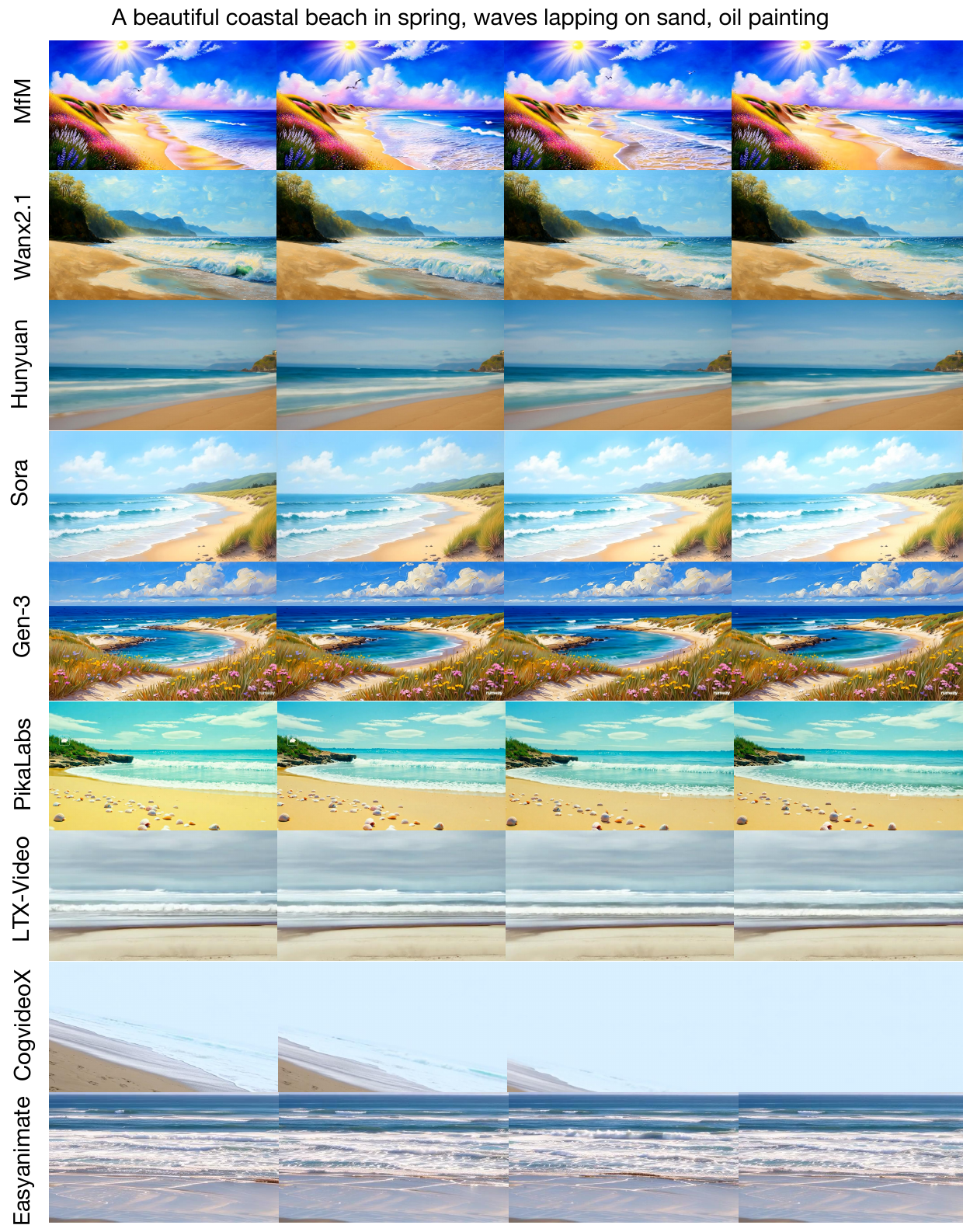}
\caption{T2V generated videos with prompt "a beautiful coastal in spring, waved lapping on sand, oil painting".}
\centering
\label{fig:t2vcase1}
\vspace{-6mm}
\end{figure}

\begin{figure}[t]
\includegraphics[width=0.96\textwidth]{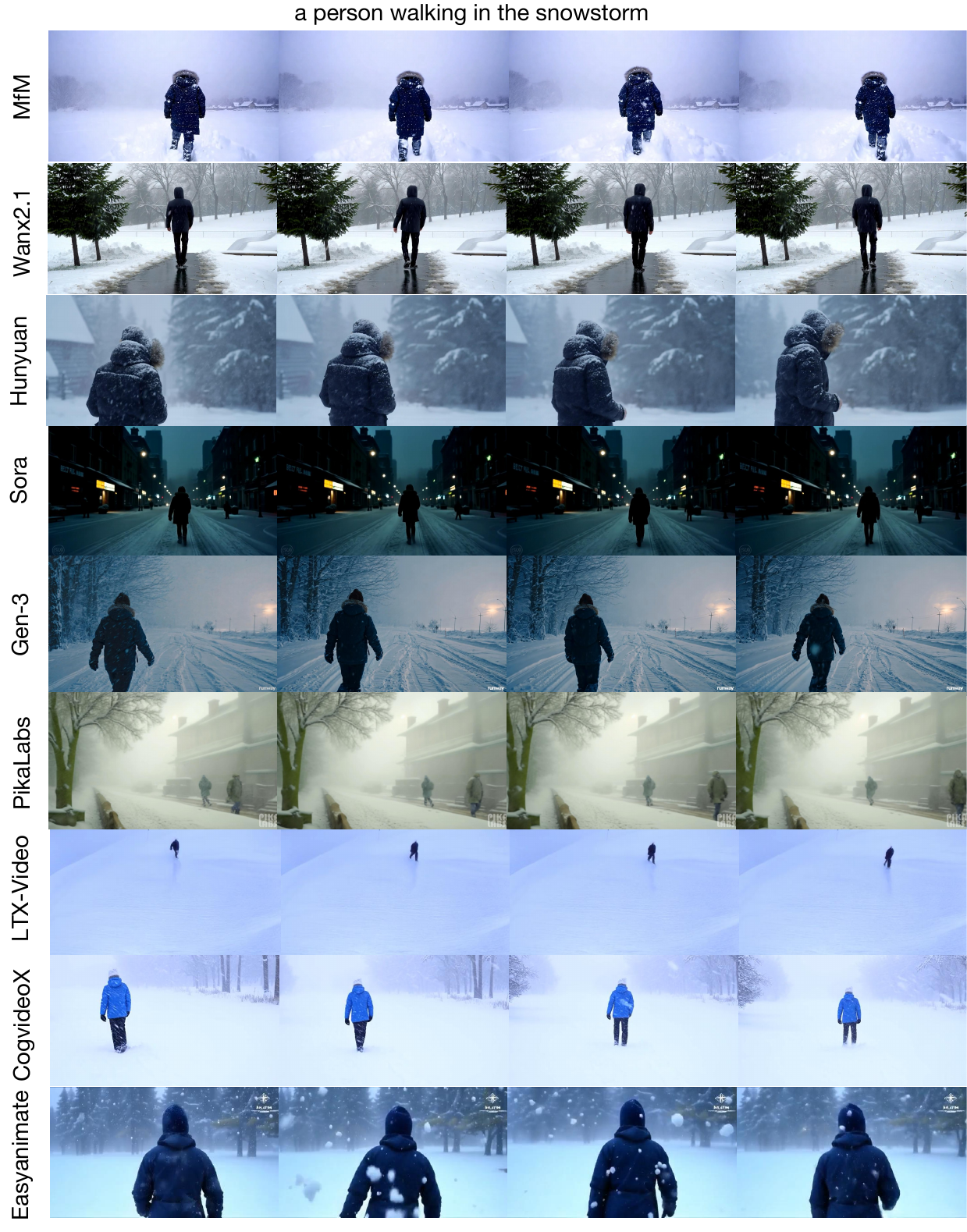}
\caption{T2V generated videos with prompt "a person walking in the snowstorm".}
\centering
\label{fig:t2vcase2}
\vspace{-6mm}
\end{figure}

\begin{figure}[t]
\includegraphics[width=0.96\textwidth]{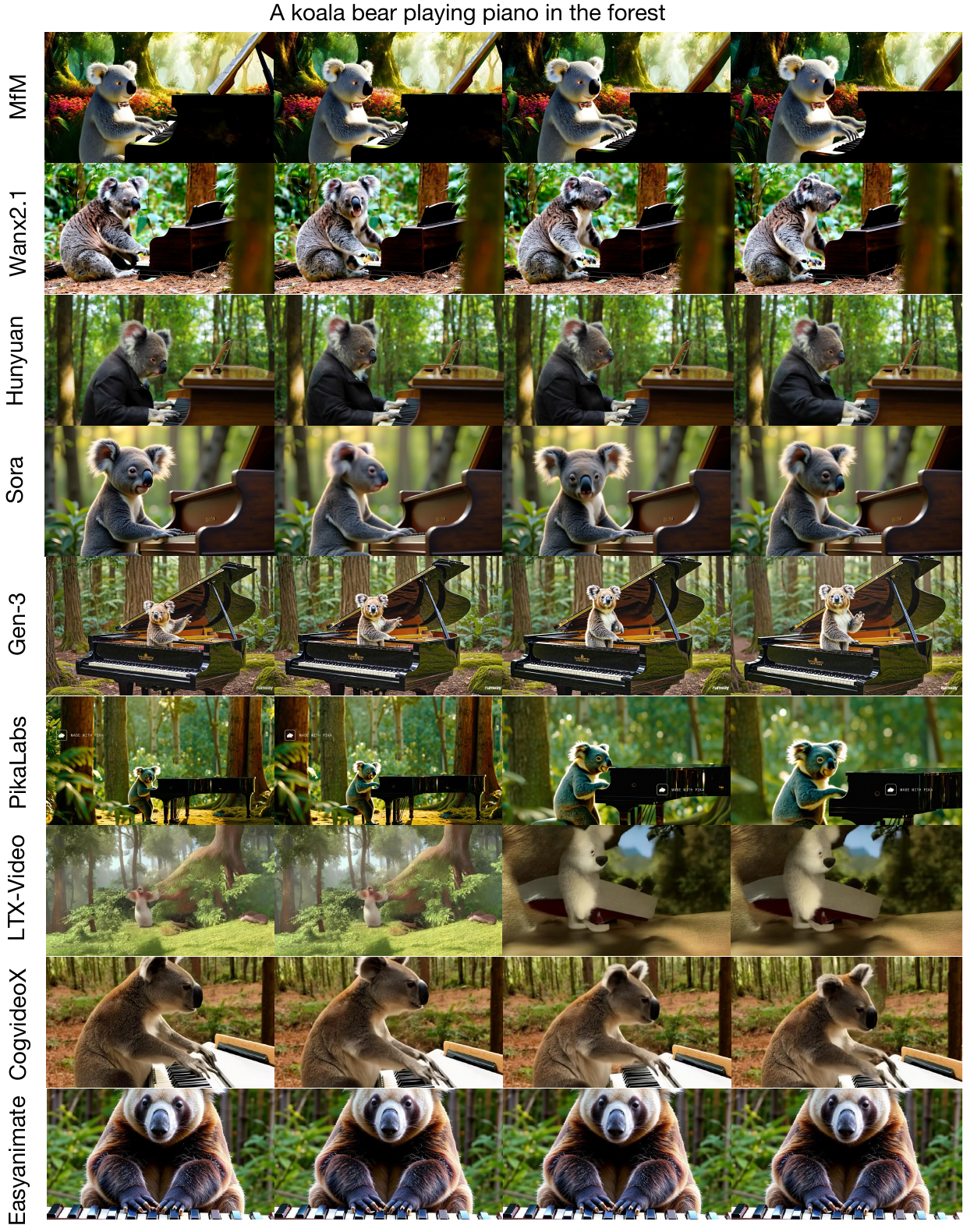}
\caption{T2V generated videos with prompt "a koala bear playing piano in the forest".}
\centering
\label{fig:t2vcase3}
\vspace{-6mm}
\end{figure}

\begin{figure}[t]
\includegraphics[width=0.96\textwidth]{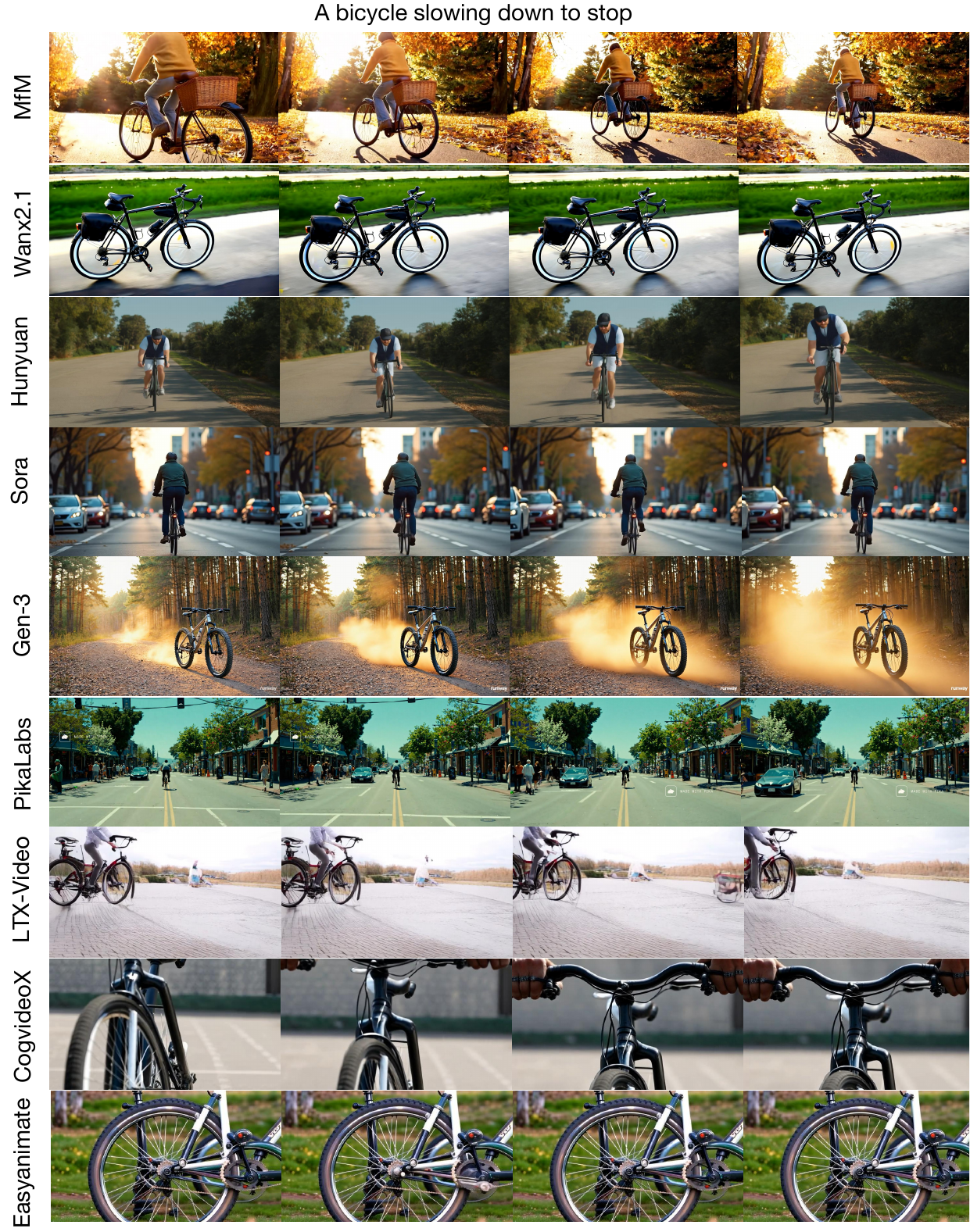}
\caption{T2V generated videos with prompt "a bicycle slowing down to stop".}
\centering
\label{fig:t2vcase4}
\vspace{-6mm}
\end{figure}

Our comprehensive qualitative analysis spans four diverse text-to-video generation scenarios—coastal beach oil painting with waves, person walking in snowstorm, koala playing piano in forest, and bicycle slowing down—as illustrated in Figures \ref{fig:t2vcase1}, \ref{fig:t2vcase2}, \ref{fig:t2vcase3} and \ref{fig:t2vcase4}. We compare MfM with Wan2.1~\cite{wan2.1}, Hunyuan~\cite{kong2024hunyuanvideo}, Sora~\cite{openai2024sora}, Gen-3~\cite{runway}, PikaLabs~\cite{pika}, LTX-Video~\cite{hacohen2024ltxvideo}, CogVideoX1.5~\cite{yang2024cogvideox}, and EasyAnimate~\cite{xu2024easyanimate}. 
From these figures and the demo videos in our provided anonymous website, we can have the following observations. 

Wanx2.1 exhibits prompt comprehension failures across multiple dimensions. For example, it fails to capture motion elements—generating a moving bicycle in Figure \ref{fig:t2vcase4} that shows no deceleration, and even producing backwards walking motion in Figure \ref{fig:t2vcase2}, contradicting natural human movement. Hunyuan produces near-identical frames in the beach scene (Figure \ref{fig:t2vcase1}), where waves show negligible movement. Additionally, Hunyuan demonstrates limited stylistic interpretation capability, completely missing the oil painting aesthetic in Figure \ref{fig:t2vcase1}. Sora produces significant contextual mismatches in several scenarios. Notably, it generates an urban nighttime scene instead of a snowstorm in Figure \ref{fig:t2vcase2}. While Sora delivers reasonable visual quality, it frequently produces minimal frame-to-frame progression, which is particularly evident in the bicycle sequence, where speed reduction is barely perceptible. 
Gen-3 generally provides good visual quality but struggles with specific prompt elements. It fails to accurately render koala coloration in Figure \ref{fig:t2vcase3}, producing an unnatural scenario where the koala is in the piano. In Figure \ref{fig:t2vcase4}, it shows a riderless bike with non-diminishing dust effects that physically contradict the slowing action specified in the prompt. 

PikaLabs demonstrates framing issues across multiple scenarios. In Figure \ref{fig:t2vcase2}, the human subject appears too small to effectively convey walking motion. This problem is even more pronounced in Figure \ref{fig:t2vcase4}, where an inappropriately wide urban composition makes the bicycle barely visible. LTX-Video exhibits the most severe quality limitations, consistently delivering washed-out, minimalist renderings across all scenarios. Most problematically, LTX-Video demonstrates dramatic mid-sequence discontinuities in Figure \ref{fig:t2vcase3}, completely changing the scene halfway through. CogVideoX generates video with small motion changes and cannot adapt to the style prompting (Figure \ref{fig:t2vcase1}). Easyaimate completely misidentifies the requested animal in Figure \ref{fig:t2vcase3}, rendering a panda instead of a koala. In Figure \ref{fig:t2vcase4}, it shows an inappropriate close-up framing of a stationary bicycle wheel, making the slowing action impossible to perceive. In contrast, MfM demonstrates superior results across all scenarios, achieving an ideal balance of prompt fidelity, motion physics, and visual quality. 

\section{Visual results of I2V generation on VBench}

\begin{figure}[t]
\includegraphics[width=0.96\textwidth]{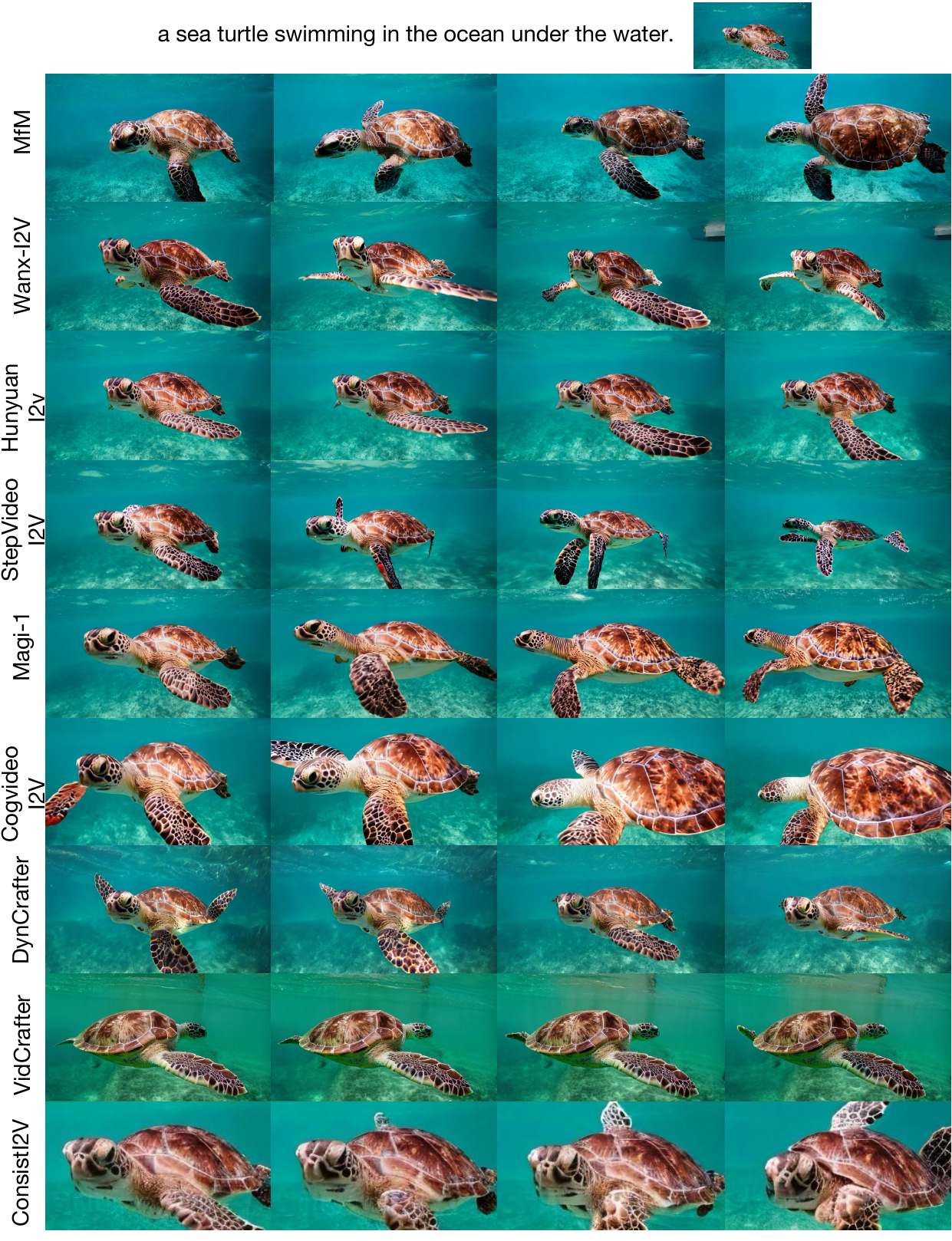}
\caption{I2V generated videos with prompt "a sea turtle swimming in the ocean under the water".}
\centering
\label{fig:i2vcase1}
\vspace{-6mm}
\end{figure}

\begin{figure}[t]
\includegraphics[width=0.96\textwidth]{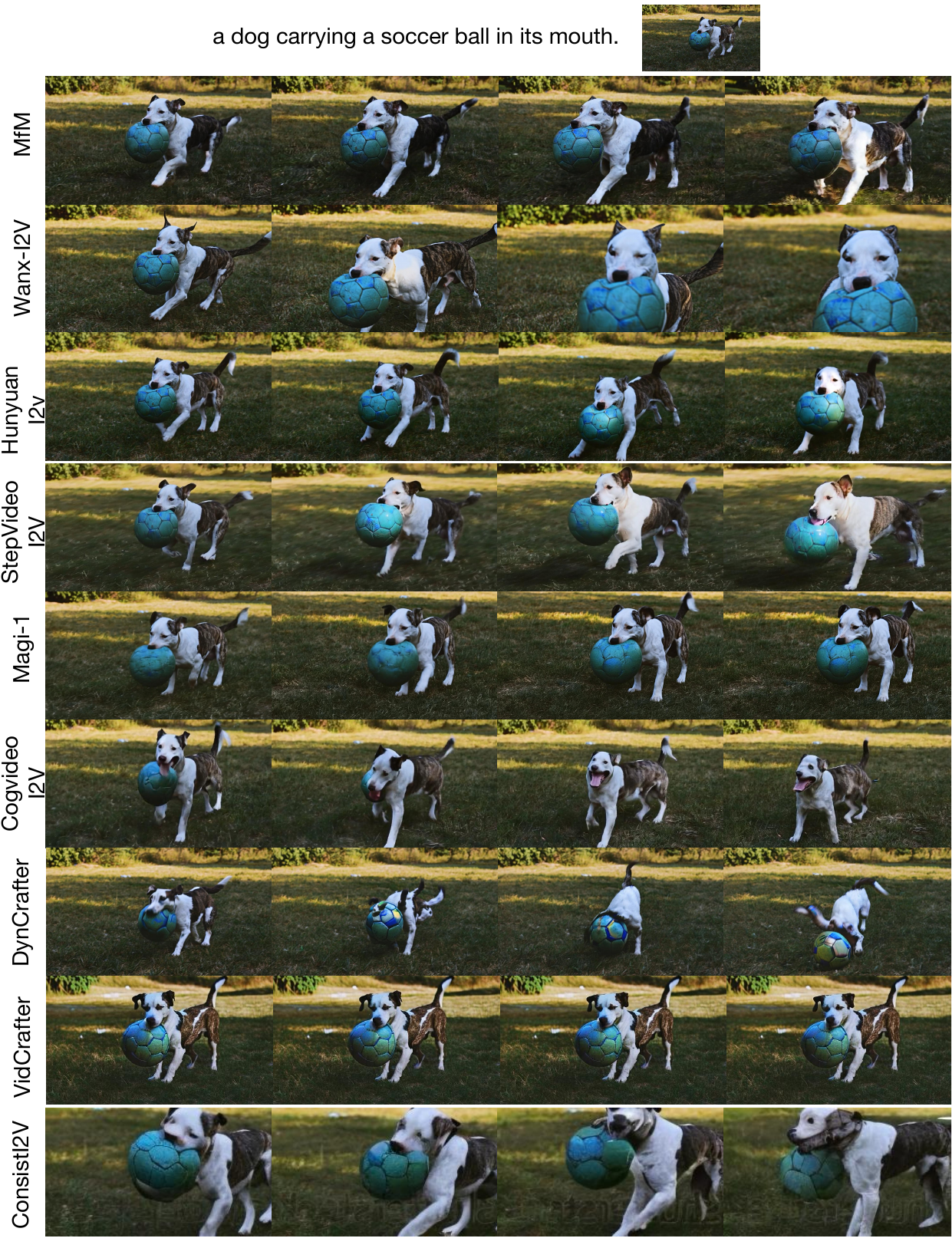}
\caption{I2V generated videos with prompt "a dog carrying a soccer ball in its mouth".}
\centering
\label{fig:i2vcase2}
\vspace{-6mm}
\end{figure}

\begin{figure}[t]
\includegraphics[width=0.96\textwidth]{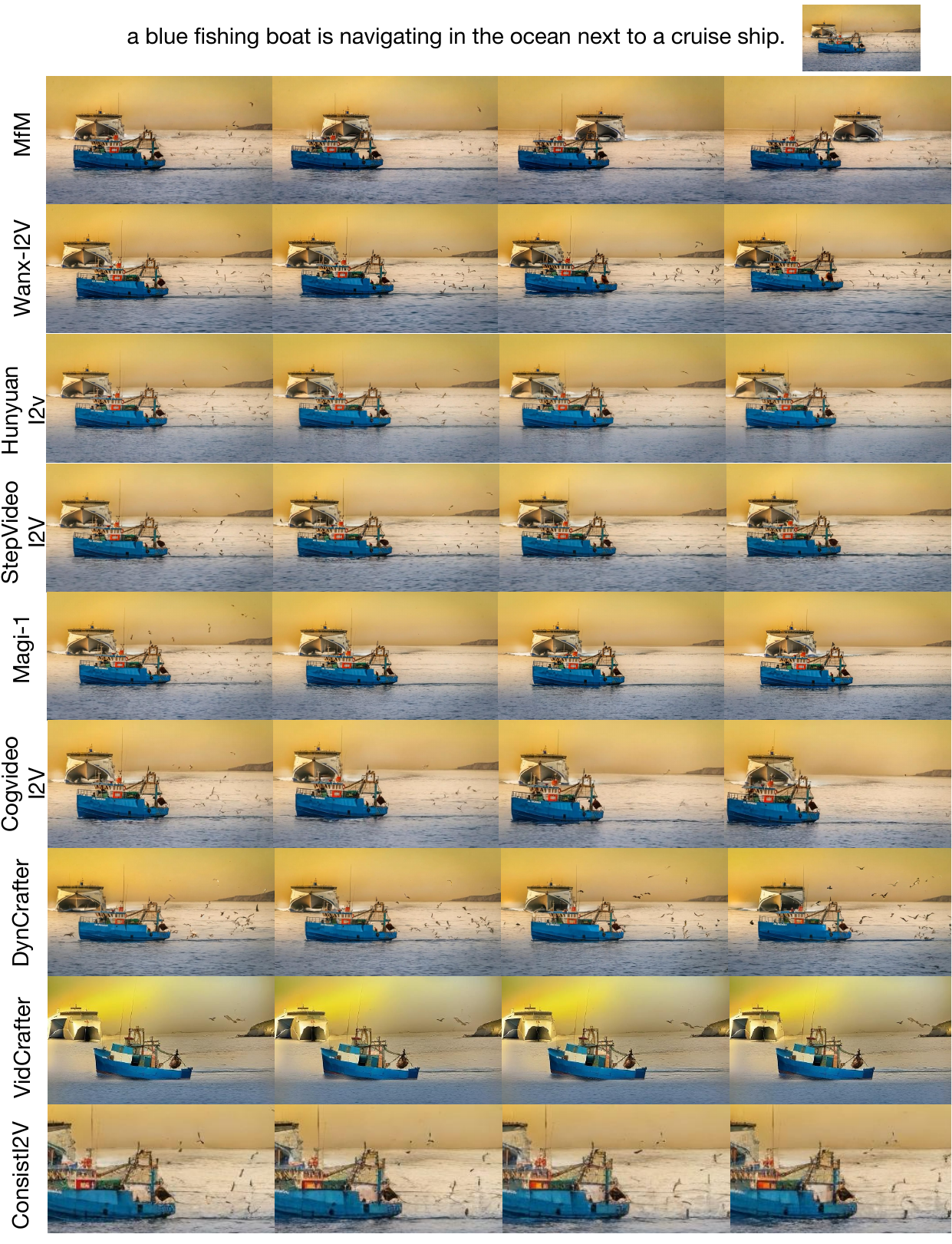}
\caption{I2V generated videos with prompt "a blue fishing boat is navigating in the ocean next to a cruise ship".}
\centering
\label{fig:i2vcase3}
\vspace{-6mm}
\end{figure}

Our qualitative analysis spans four diverse cases—swimming turtle, dog carrying a soccer ball, fishing boat navigation, and galloping horses—as illustrated in Figures \ref{fig:i2vcase1}, \ref{fig:i2vcase2}, \ref{fig:i2vcase3}. These scenarios were selected to evaluate model performance across a spectrum of challenges, including animal locomotion, object interaction, environmental dynamics, and atmospheric conditions. We compare our MfM with Wanx-I2V~\cite{wan2.1}, Hunyuan-I2V~\cite{kong2024hunyuanvideo}, Magi-1~\cite{magi1}, Step-Video~\cite{ma2025stepvideot2v}, DynamicCrafter~\cite{xing2023dynamicrafter}, VideoCrafter-I2V~\cite{chen2024videocrafter2}, I2VGen-XL~\cite{zhang2023i2vgen}, CogvideoX-I2V~\cite{yang2024cogvideox}, ConsistI2V~\cite{ren2024consisti2v}.
From these figures and the demo videos in our provided anonymous website, we can have the following observations. 

Hunyuan-I2V demonstrates minimal temporal progression across all scenarios, producing sequences with negligible motion variation. This is particularly evident in the turtle (Figure \ref{fig:i2vcase1}) and fishing boat (Figure \ref{fig:i2vcase3}) examples. Furthermore, Hunyuan-I2V introduces anatomical inconsistencies in the horse sequence, rendering equine subjects with only three legs in later frames—a critical biological implausibility. Wanx-I2V can produce reasonable animal movement, but sometimes fail to capture essential action descriptors. For example, it fails to generate the "navigating" movement explicitly specified in the boat prompt (Figure \ref{fig:i2vcase3}). StepVideo-I2V suffers from visual artifacts across multiple dimensions, including anatomical anomalies (abnormal turtle fin articulation in Figure \ref{fig:i2vcase1}), subject identity inconsistencies (altered dog appearance in Figure \ref{fig:i2vcase2}), and most strikingly, fundamental scene misinterpretation in the horse sequence. CogVideo-I2V demonstrates object consistency failures, including problematic size variations in the turtle sequence and unstable object interactions in the dog example. DynamiCrafter exhibits even more pronounced temporal instability, with objects and environmental elements changing unnaturally between consecutive frames—most evident in the inconsistent appearance of soccer ball and geometric distortions of the dog subject in Figure \ref{fig:i2vcase2}. VidCrafter and ConsistI2V both struggle with maintaining prompt fidelity, frequently altering the fundamental identity characteristics in the conditioning image. This prompt deviation is particularly pronounced in the dog sequence (Figure \ref{fig:i2vcase2}), where breed characteristics, coat patterns, and contextual elements shift significantly from the reference image. In contrast, our MfM and Magi-1 achieve both robust identity preservation and convincing motion dynamics in all test cases. 

\section{User study of I2V generation}

To comprehensively evaluate MfM's effectiveness in generative tasks, we conduct a user study specifically focused on image-to-video (I2V) generation, comparing against nine I2V generation methods whose models are publicly available: Wanx-I2V~\cite{wan2.1}, Hunyuan-I2V~\cite{kong2024hunyuanvideo}, StepVideo-I2V~\cite{ma2025stepvideot2v}, Magi-1~\cite{magi1}, Cogvideo-I2V~\cite{yang2024cogvideox}, I2VGenXL~\cite{zhang2023i2vgen}, DynamiCrafter~\cite{xing2023dynamicrafter}, VideoCrafter~\cite{chen2024videocrafter2}, and ConsistI2V~\cite{ren2024consisti2v}. The user study comprised 10 test cases encompassing various content categories, including animal motion, human activities, scenic close-ups, and vehicular movement. We invited 10 participants and asked them to rank the top three generated videos for each case based on visual quality and semantic consistency. Table~\ref{tab:userstudy} presents the average Top-1 and Top-3 rates for all methods. The results clearly show that MfM outperforms all competitors, achieving a 64.29\% Top-1 rate and an 85.71\% Top-3 rate. Wanx-I2V ranks second with 20.00\% Top-1 and 71.43\% Top-3 rates, respectively. Hunyuan-I2V and StepVideo-I2V demonstrate moderate performance with Top-3 rates of approximately 40\%, despite Top-1 rates below 5\%. Notably, Magi-1, DynamiCrafter, VideoCrafter, and ConsistI2V fail to secure any Top-1 selections and exhibit minimal presence in Top-3 rankings. These results reveal MfM's superior capability in generating high-quality image-to-video content that consistently meets human evaluation criteria.

\begin{table}[tbp]
\vspace{-2mm}
\centering
\caption{User votes on 10 image-to-video generation outputs}
\small
\setlength{\tabcolsep}{4pt}  
\normalsize{
\resizebox{\textwidth}{!}{
\begin{tabular}{l|*{10}{c}}
\toprule
Method & MfM & \makecell[c]{Wanx \\ I2V} & \makecell[c]{Hunyuan\\I2V} & \makecell[c]{StepVideo\\I2V} & Magi-1 & \makecell[c]{Cogvideo\\I2V} & I2VGenXL & DynCrafter & VidCrafter & ConsistI2V \\
\midrule
Top-1 Rates & 64.29\% & 20.00\% & 1.43\% & 4.29\% & 0.00\% & 5.71\% & 4.29\% & 0.00\% & 0.00\% & 0.00\% \\
Top-3 Rates & 85.71\% & 71.43\% & 40.00\% & 41.43\% & 12.86\% & 30.00\% & 12.86\% & 0.00\% & 4.29\% & 1.43\% \\
\bottomrule
\end{tabular}
}}
\vspace{-2mm}
\label{tab:userstudy}
\end{table}

\section{Visual results of multi-task generation}

\begin{figure}[t]
\includegraphics[width=0.96\textwidth]{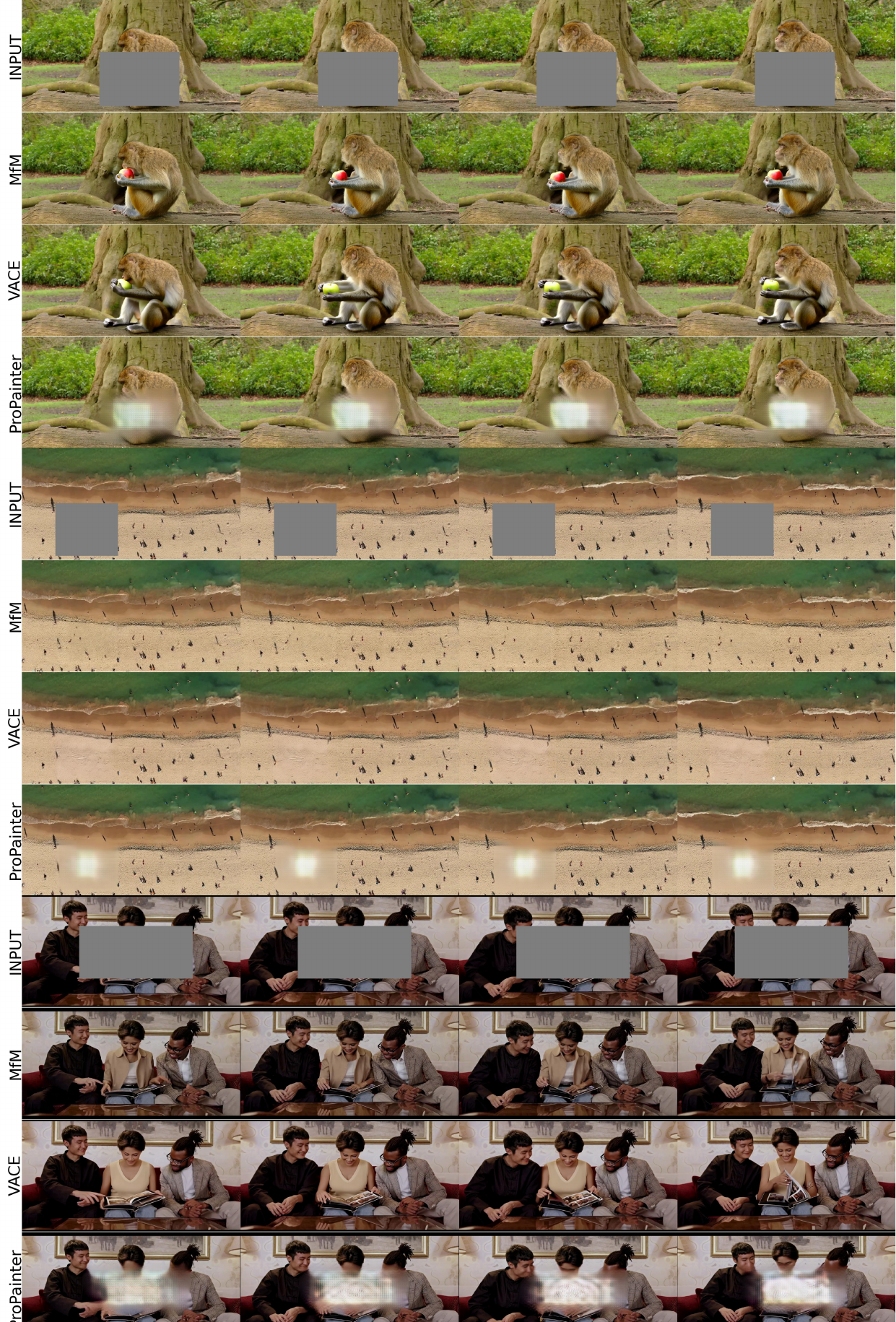}
\caption{Visual comparison on task of video inpainting.}
\centering
\label{fig:vinp}
\vspace{-6mm}
\end{figure}

\begin{figure}[t]
\includegraphics[width=0.85\textwidth]{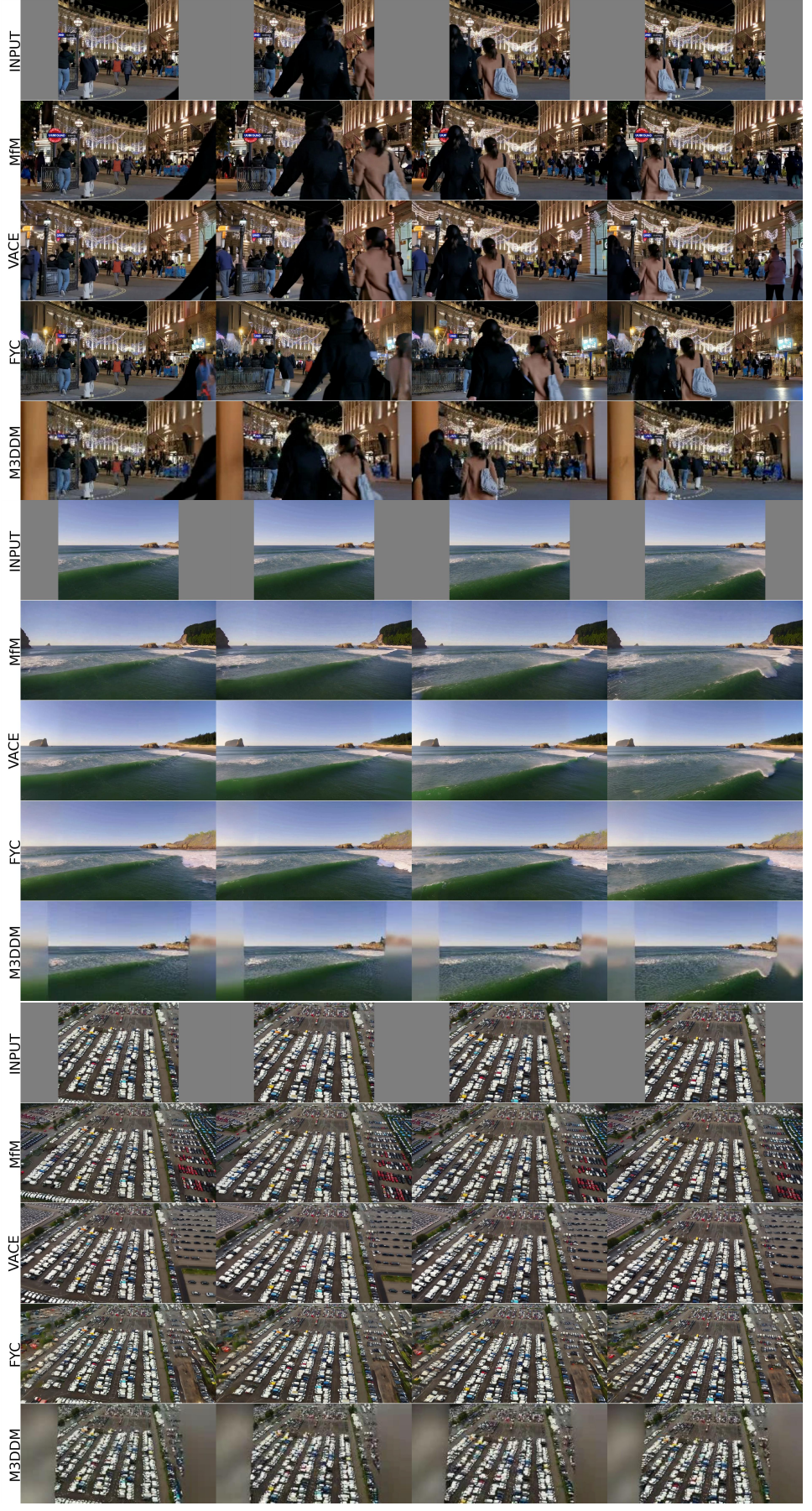}
\caption{Visual comparison on task of video outpainting.}
\centering
\label{fig:voup}
\vspace{-6mm}
\end{figure}

\begin{figure}[t]
\includegraphics[width=0.96\textwidth]{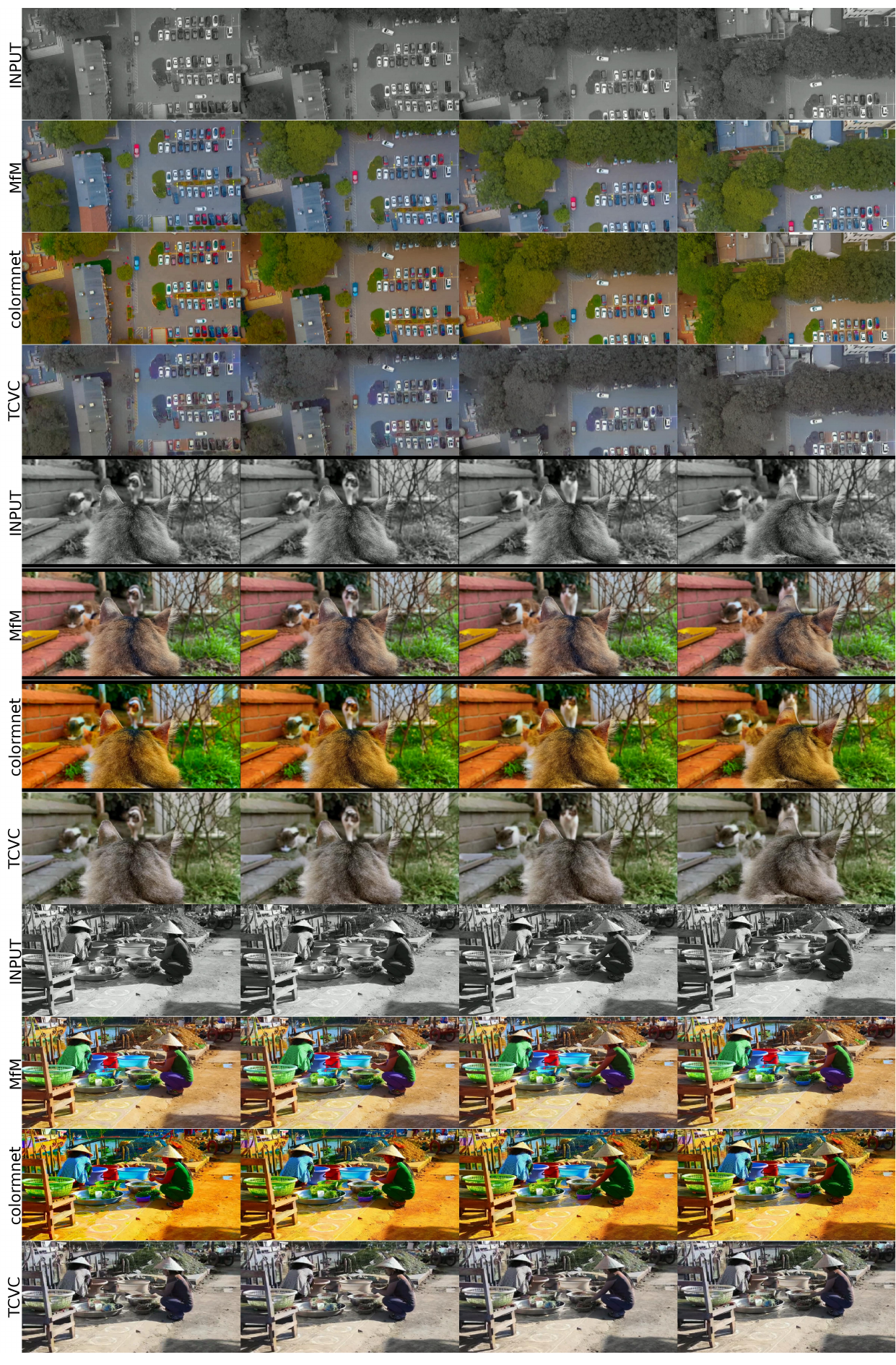}
\caption{Visual comparison on task of video colorization.}
\centering
\label{fig:vcolor}
\vspace{-6mm}
\end{figure}

\begin{figure}[t]
\includegraphics[width=0.96\textwidth]{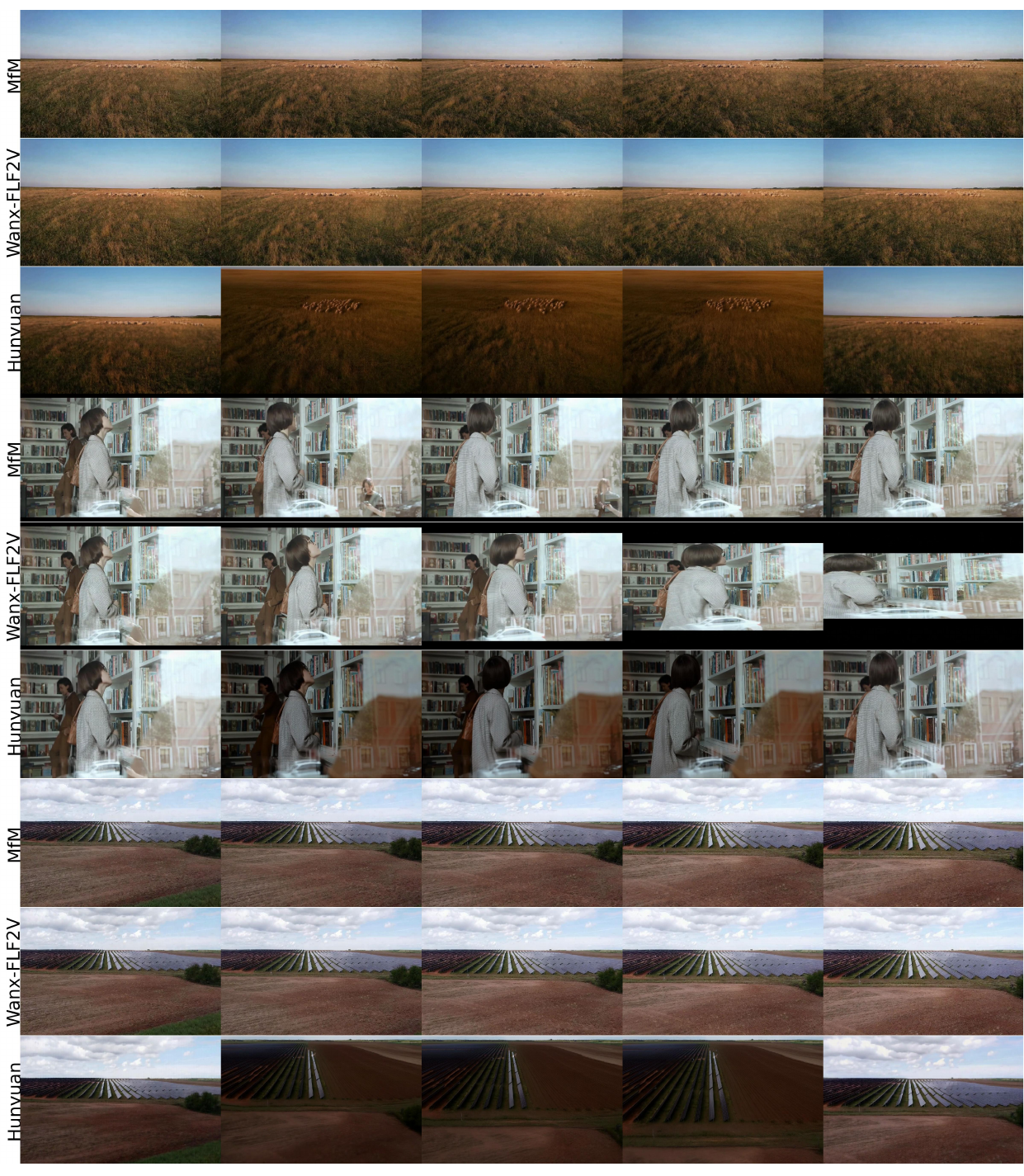}
\caption{Visual comparison on task of first-last-frame to video.}
\centering
\label{fig:FLF2V}
\vspace{-6mm}
\end{figure}

The visual results of four representative video tasks, including VINP, VOUTP, VCOLOR and FLF2V,  are illustrated in Figure~\ref{fig:vinp}, Figure~\ref{fig:voup}, Figure~\ref{fig:vcolor}, Figure~\ref{fig:FLF2V}, respectively. 

For VINP, we compare our MfM with VACE~\cite{jiang2025vace} and ProPainter~\cite{zhou2023propainter} across three diverse scenarios in Figure~\ref{fig:vinp}. We see that MfM demonstrates superior performance in maintaining visual fidelity and temporal consistency. Specifically, both MfM and VACE produce reasonably coherent results in the first and third cases, where they successfully reconstruct the masked region with detail preservation and natural integration with the surrounding environment. However, in the second case, VACE shows inconsistencies in intensity distribution and color. ProPainter exhibits severe blurring and artifacts in the inpainted region, failing to properly reconstruct the subject and completely losing the details.

For VOUTP, we compare MfM with VACE~\cite{jiang2025vace}, Follow-Your-Canvas (FYC)~\cite{chen2024follow}, and M3DDM~\cite{fan2023M3DDm}. The comparisons on three diverse scenarios are illustrated in Figure~\ref{fig:voup}. MfM demonstrates exceptional consistency and contextual understanding across all test cases. VACE shows moderate capabilities but with noticeable limitations. While it produces acceptable wave continuation in the ocean scene, it generates noticeable brightness mismatches between the original and generated regions in the second case. FYC suffers from the brightness mismatches in the second case; what's more, it fails to complete the leg of the person in the first frame of the first case. M3DDM exhibits significant limitations in this task. It generates blurred outputs and visually jarring discontinuities around the generated areas.

For VCOLOR, we compare MfM with Colormnet~\cite{yang2024colormnet} and TCVC\cite{zhang2023TCVC} in Figure~\ref{fig:vcolor}. We see that MfM demonstrates excellent contextual understanding performance and maintains superior temporal color stability between adjacent frames. Compared with other baselines, it achieves more complete colorization coverage without introducing grayscale artifacts while preserving realistic lighting conditions. Colormnet shows reasonable performance on these cases but suffers from saturation issues in the last two cases. TCVC exhibits substantial limitations across all test scenarios, with large portions remaining in grayscale and the overall color tone appearing excessively dull and lifeless.

For FLF2V, we compare MfM with Wanx-FLF2V~\cite{wan2.1} and Hunyuan (with keyframe LoRA)~\cite{kong2024hunyuanvideo}. The comparisons on three scenarios are illustrated in Figure~\ref{fig:FLF2V}. We see that both MfM and Wanx-FLF2V deliver natural motion interpolation between the first frame and the last frame without jarring transitions, as shown in the first and third cases. But Wanx-FLF2V performs abnormally in the second case, where the video frames are unexpectedly compressed vertically at the end, altering the aspect ratio. Hunyuan exhibits severe limitations in the first and third cases. It produces intermediate frames with a different viewpoint and a noticeably darkened color tone, resulting in jarring visual transitions.

\section{Visual results of ablation study}

\begin{figure}[t]
\includegraphics[width=0.96\textwidth]{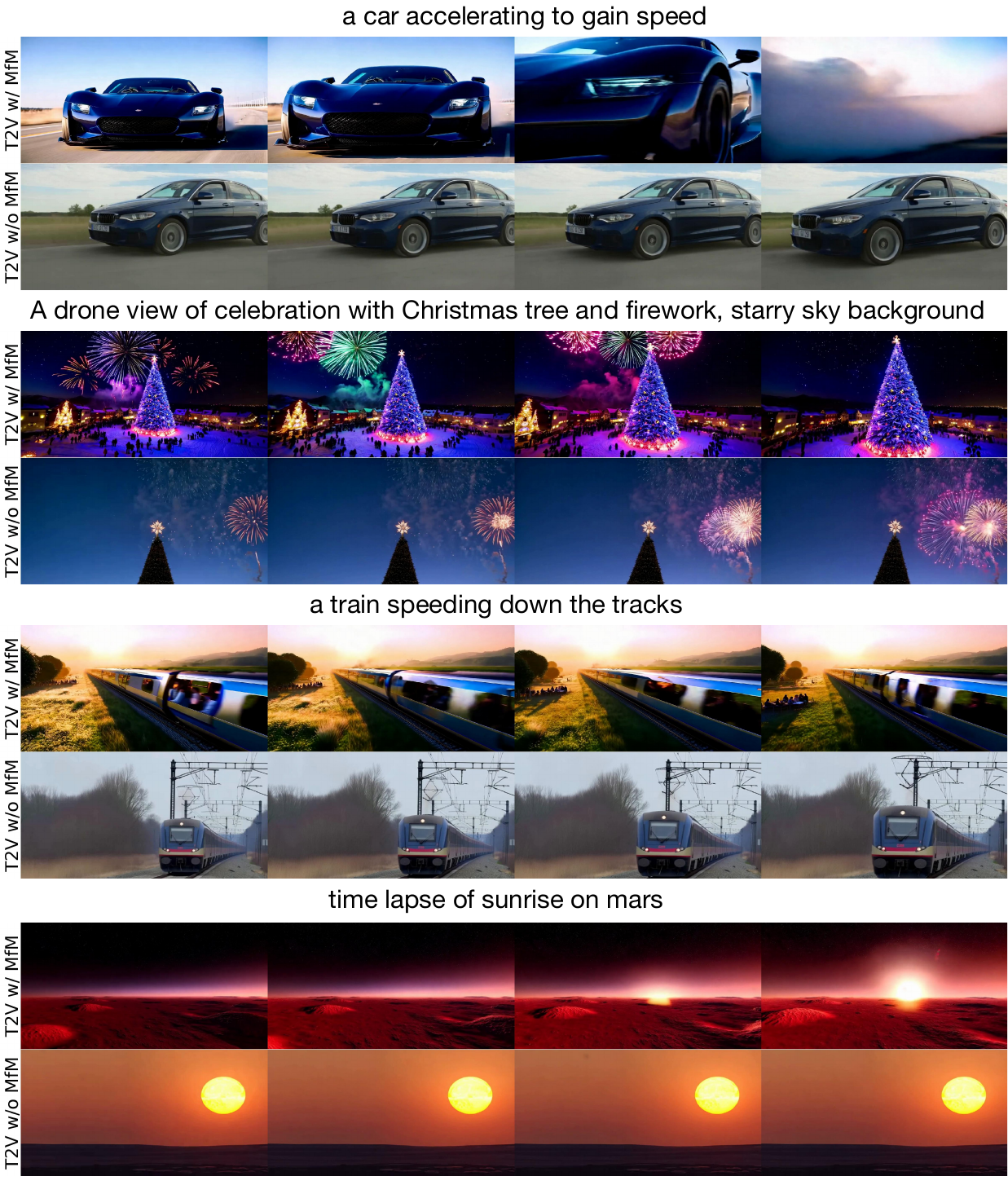}
\caption{Ablation study on t2v task using MfM with/without multi-task training.}
\centering
\label{fig:ablation}
\vspace{-6mm}
\end{figure}

The ablation study results on four scenarios are illustrated in Figure~\ref{fig:ablation}, which provides visual evidence to support that multi-task training significantly improves the temporal dynamics of the generated videos. We can see that T2V w/ MfM demonstrates cinematographic qualities, including smooth and flexible camera movements, as well as vivid and evolving patterns. For instance, in the first case, the varying angle of the video effectively captures the dynamic essence of 'gain speed'; in the second case, dynamic color transitions and the natural progression of pyrotechnic effects are illustrated; in the third case, the train is portrayed with appropriate motion blur; and in the fourth, the celestial progression is dramatically captured, with the sun emerging and intensifying across the horizon, accompanied by corresponding atmospheric lighting changes. In contrast, T2V w/o MfM exhibits minimal camera movement, with limited perspective variation and an almost static side view throughout the sequence. Furthermore, in the last case, T2V w/o MfM produces nearly identical frames of a static sun, with little temporal progression.

\section{Failure cases}
\begin{figure}[t]
\includegraphics[width=0.96\textwidth]{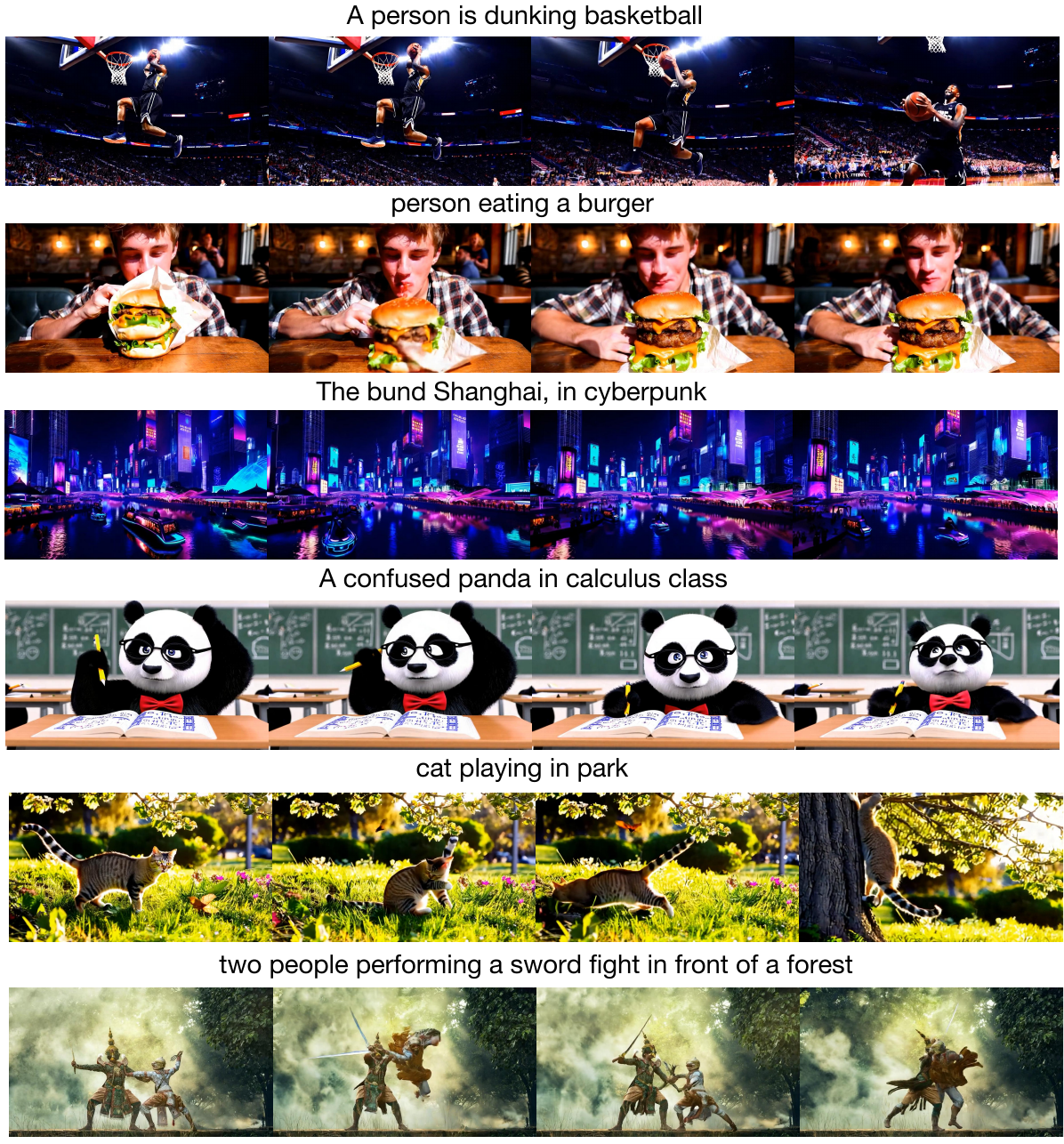}
\caption{Failure cases of our MfM.}
\centering
\label{fig:failure}
\vspace{-6mm}
\end{figure}

While MfM demonstrates strong performance across T2V and I2V generation tasks, it also occasionally produces failure cases, as illustrated in Figure~\ref{fig:failure}. First, in complex interaction scenarios, MfM may produce physically implausible object relationships. For instance, in the basketball dunking sequence (first row), the ball incorrectly traverses the net rather than entering the basket properly. Similarly, in the burger eating sequence (second row), the burger wrapper abruptly merges into the burger in the intermediate frames. Second, MfM also exhibits limitations in generating videos that contain words; this is particularly evident in the cyberpunk cityscape (third row) and the animated panda scene (fourth row). Finally, for sequences involving rapid motion, we observe temporal artifacts manifesting as duplicated or misplaced features. This is exemplified in the cat playing sequence (fifth row), where an anomalous second tail-like object appears near the cat's head in intermediate frames. Meanwhile, in the sword fighting sequence (last row), the character on the right undergoes noticeable variations and distortions in the intermediate frames. Future work will be conducted to further improve the performance of MfM on these scenarios.

\section{Broader impacts and safeguards}

Our proposed MfM approach demonstrates significant advancements in video generation capabilities across multiple tasks. While these improvements offer substantial benefits for creative industries, and media production, we acknowledge several potential societal impacts that require careful consideration. On the positive side, MfM can democratize high-quality video content creation, enhance the viewing experience of historical videos through color restoration, and reduce production costs for educational and entertainment media. However, like all generative video technologies, MfM could potentially be misused to create misleading or deceptive content, including generating videos that might impact privacy, security, or public discourse.

To address these concerns, we have implemented several safeguards throughout our development process. First, our training datasets were carefully curated to exclude problematic content and ensure diverse representation. Second, we will integrate fingerprinting capabilities into our model to add fingerprints into the generated videos, helping identify AI-generated content.

\end{document}